\documentclass{article} 
\usepackage{iclr2027_conference,times}

\iclrfinalcopy 

\usepackage{amsmath,amsfonts,bm}

\def\eqref#1{equation~\ref{#1}}

\def\1{\bm{1}}

\DeclareMathAlphabet{\mathsfit}{\encodingdefault}{\sfdefault}{m}{sl}
\SetMathAlphabet{\mathsfit}{bold}{\encodingdefault}{\sfdefault}{bx}{n}

\usepackage{hyperref}
\usepackage{url}

\usepackage[utf8]{inputenc} 
\usepackage[T1]{fontenc}    
\usepackage{hyperref}       
\usepackage{url}            
\usepackage{booktabs}       
\usepackage{amsfonts}       
\usepackage{nicefrac}       
\usepackage{microtype}      
\usepackage{xcolor}         

\usepackage{algorithm}
\usepackage{algorithmic}
\usepackage[most]{tcolorbox}
\usepackage{subcaption}
\usepackage{tabularray}
\usepackage{wrapfig}
\usepackage{multirow}
\usepackage{graphicx}
\usepackage{makecell}
\usepackage{fontawesome}
\usepackage{enumitem}

\usepackage{amsmath}
\usepackage{amssymb}
\usepackage{mathtools}
\usepackage{amsthm}
\usepackage{amsfonts}
\usepackage{bbm}

\newtheorem{proposition}{Proposition}

\title{Revealing what benchmark scores hide: Jointly measuring the breadth and \mbox{reliability} of LLM capabilities using accessibility \mbox{profiles}}

\author{Kazem Faghih\thanks{Correspondence to: \texttt{<kazemf@umd.edu>}.} \quad
  Yize Cheng \quad
  Shoumik Saha \\
  \textbf{Mobina Pournemat \quad Armin Gerami \quad Soheil Feizi} \\
  Department of Computer Science, University of Maryland, College Park \\
  \faGithub~\textbf{Code \& data}: \url{https://github.com/kazemf78/access-eval}
}

\begin{document}

\maketitle
\lhead{Preprint} 

\begin{abstract}
Language model benchmarks assign each question a single outcome from one fixed wording, even though the same underlying request can be expressed in many equivalent ways. We argue that the quantity of interest should therefore not be correctness on one wording, but the \emph{accessibility} of an inquiry, which we define as the probability that a model succeeds across meaning-preserving formulations. The distribution of accessibility across inquiries reveals a structure that accuracy and existing best-, worst-, or average-case prompt scores collapse: how broadly a model's capabilities extend across inquiries and how dependably those capabilities can be elicited. Two models can have the same average performance while one succeeds across more inquiries unreliably and the other succeeds across fewer inquiries but more consistently, leading to opposite preferences depending on whether the task allows multiple retries until one success or when repeated consistent success is required. Across six benchmarks and more than twenty models, this tradeoff is common: for about half of the model pairs with nearly equal mean accessibility, which model looks better depends on whether one success across five formulations is enough or all five must succeed, quantities we call Access@5 and Reliable@5, respectively. For some pairs the two measures diverge significantly, by up to 10 points in opposite directions. Accessibility also reveals that scaling, reasoning, and prompt optimization can redistribute capability across inquiries in ways aggregate accuracy obscures, sometimes broadening the set of inquiries a model can succeed on while making success on them less repeatable. Accessibility turns formulation sensitivity from a nuisance into a measurable property of capability, distinguishing what a model can sometimes succeed on from what users can actually depend on.
\end{abstract}

\section{Introduction}
\label{sec:intro}

Language model benchmarks ask each question in one fixed wording, but
people who use a model in reality can phrase the same request in many different ways. A model
that answers one wording correctly may fail on another, yet a benchmark score only
tells us how the model handled the benchmark's wording, not how dependably it
handles the underlying request.

It is well established that models are sensitive to wording: rephrasing
instructions, templates, or questions without changing their meaning can change
both scores and model rankings
\citep{mizrahi2024state,cao2024worst,habba2025dove,faghih2026same}. In response,
recent work evaluates models over several wordings and reports the best, worst,
or average score across them
\citep{mizrahi2024state,cao2024worst,polo2024prompteval,lior2025reliableeval}.
These summaries establish that wording matters, but they do not show how success is distributed across the underlying questions. This distinction is important because the same average performance can arise from very different model behaviors. A model may succeed on a relatively narrow set of questions but answer those questions consistently however they are phrased. Another may succeed on a much broader set of questions, but only under specific formulations. The first offers narrower but dependable capability; the second broader but less dependable capability. A single benchmark score, and even an average over multiple wordings, can make these models look similar despite this fundamental difference.

We therefore treat each \textbf{underlying question}, which we call an
\emph{\textbf{inquiry}}, as the object of evaluation, rather than any one wording of it, which we call a
\emph{\textbf{formulation}} of that inquiry. For each inquiry, we measure its \emph{\textbf{accessibility}}: the
share of meaning-preserving formulations that the model answers correctly. We
generate the formulations automatically and keep only those that independent
judges verify as equivalent to the original. Human annotators agree with this
verification on 97--98\% of annotated pairs. 

On six benchmarks spanning factual question
answering, mathematical reasoning, general knowledge, and tool use, and by evaluating more than twenty models, we show that the spread of accessibility across a
benchmark's inquiries, which we call the model's \emph{\textbf{accessibility profile}}, answers several
practical questions at once: how often the model succeeds on a randomly worded
request, which inquiries it answers correctly under every formulation, which
under at least one, and how it performs when a user can retry versus when every
attempt must succeed.


\paragraph{Contributions and key findings.}

\begin{enumerate}[leftmargin=*]

\item \textbf{Inquiry-level accessibility.}
We introduce inquiry-level accessibility evaluation, where an accessibility profile simultaneously
captures average performance, capability that is reachable under at least some
formulations, and capability that remains reliable across formulations. It also
determines performance when one success across repeated attempts is enough
versus when every attempt must succeed. We further theoretically characterize when these two
views necessarily prefer different models
(Proposition~\ref{prop:single-crossing}).

\item \textbf{RQ1: What does accessibility reveal that accuracy hides?}
Models with similar average performance can differ sharply in how widely their
successes extend and how consistently they hold. For about half of the model
pairs with nearly equal mean accessibility, which model looks better depends on whether one success across five formulations is enough or all five
must succeed, and for some pairs the gap is significant, up to 10 points in each
direction.

\item \textbf{RQ2: Is this variation caused by the formulation or by decoding
noise?} In a controlled study, answers to different equivalent formulations
disagree in correctness more often than repeated samples of one formulation in
11 of 12 settings, and varying the formulation raises the chance of at least one
success while lowering the chance that all attempts succeed.

\item \textbf{RQ3: What does accessibility reveal about model changes?}
Scaling, reasoning, and prompt optimization redistribute capability across
inquiries in ways aggregate scores conceal. For example, enabling native thinking in Qwen3
changes MMLU accuracy by only $-3.0$ points on average, yet makes 25\% of
inquiries more accessible and 32\% less, and lowers the share answered reliably
across formulations by 10.9 points.

\end{enumerate}

\section{Related Work}
\label{sec:related}

\paragraph{Prompt sensitivity and multi-prompt evaluation.}
Language models are sensitive to meaning-preserving changes in instruction
templates, formatting, and wording
\citep{sclar2024formatspread,chatterjee2024posix,zhuo2024prosa,habba2025dove}, and
paraphrases of benchmark questions change individual answers even when aggregate
accuracy and coarse rankings are stable
\citep{elazar2021pararel,faghih2026same,robustness2025benchmark,roparq2025}.
Evaluation over multiple prompts has been proposed in response:
use-case-specific metrics over instruction paraphrases \citep{mizrahi2024state},
worst-, best-, and average-case performance over paraphrases of each user query
\citep{cao2024worst}, efficient estimation of performance quantiles across
templates \citep{polo2024prompteval}, and the number of prompt resamplings needed
for a stable estimate \citep{lior2025reliableeval}; related work constructs
user-grounded paraphrases \citep{augment2025}, removes surface-form artifacts in
scoring \citep{paraeval2026,hua2025flaw}, or trains for paraphrase invariance
\citep{roparq2025}. These works primarily summarize prompt-level or aggregate
variation. We instead take the success probability of each inquiry across its
formulations as the primitive object, study the distribution of these
probabilities across inquiries, derive operational quantities from that one
distribution, and analyze how it moves under model interventions.

\paragraph{Repeated trials and model changes.}
pass@$k$ \citep{chen2021codex} and pass$^k$ \citep{yao2024taubench} measure
whether at least one or all of $k$ attempts succeed, usually over decoding
samples, and reliable@$k$ has been defined over $k$ related prompts
\citep{ifevalpp2026}; perturbation-based uncertainty estimation varies inputs as
well as samples \citep{gao2024spuq,cox2025mapping}. We derive the at-least-once
and all-succeed quantities from the accessibility profile, give a condition under
which they rank models in opposite orders, and compare formulation with decoding
variability in terms of task success. Model updates are known to break inputs
that previously worked \citep{bansal2019backward,yan2021pct,echterhoff2024muscle},
and studies of scaling, reasoning, and prompt optimization usually report changes
in aggregate accuracy \citep{gepa2025,miprov2_2024}. We analyze such changes
through each inquiry's accessibility before and after, which exposes movement and
regressions that leave the canonical label unchanged.

\section{Inquiry-Centric Evaluation}
\label{sec:inquiry-evaluation}

\subsection{Inquiries, Formulations, and Accessibility}
\label{sec:inquiries-formulations}
\label{sec:inquiry-accessibility}

A benchmark item is one textual realization of an underlying question. We call
the underlying semantic object an \emph{inquiry} $q$ and each meaning-preserving
realization of it a \emph{formulation} $f$ (e.g., \emph{"What is the capital of
France?"} and \emph{"Which city is France's capital?"} are both valid formulations of the same inquiry). We hold the deployment
configuration $c$ fixed (system and task instructions, tools, decoding, and
scoring) and vary only the formulation. For a model $M$ and a distribution $D_q$
over reasonable formulations of $q$, the \emph{accessibility} of $q$ is defined as:
\begin{equation}
A_q(M;c)=\Pr_{f\sim D_q}\bigl[Y(M,c,f)=1\bigr],
\label{eq:accessibility}
\end{equation}
where $Y(M,c,f)$ indicates that the desired behavior (a correct answer, a solved
problem, a correct tool call) is observed for formulation $f$. When $M$ and $c$ are fixed, we write $A_q$ for brevity.

\subsection{Estimating Accessibility}
\label{sec:estimating-accessibility}

$D_q$ is generally unknown, so we approximate it with a uniformly weighted set of
validated formulations $f_1,\dots,f_m$ 
and estimate
\begin{equation}
\hat A_q(M;c)=\frac1m\sum_{i=1}^m y_i ,
\label{eq:accessibility-estimate}
\end{equation}
where $y_i$ indicates success on $f_i$. Paraphrasing is our procedure for
sampling formulations. This estimate requires
formulations that preserve the inquiry while covering meaningful variation
(Section~\ref{sec:formulation-generation}).

\subsection{Accessibility Profiles}
\label{sec:accessibility-profiles}

Over a benchmark $B$ with inquiries $q_1,\dots,q_n$, a model induces an
\emph{accessibility profile} $\{A_{q_i}(M;c)\}_{i=1}^n$, estimated by
$\{\hat A_{q_i}(M;c)\}$. For interpretation, we call an inquiry \emph{absent}
if $\hat A=0$, \emph{brittle} if $0<\hat A<1$, or \emph{reliable} if $\hat A=1$, and
\emph{latent} as long as $\hat A>0$. 
Because accessibility is estimated from finitely many formulations, these
labels are empirical: \emph{reliable} means success on every formulation we
tested, and \emph{absent} means failure on every formulation we tested, not
necessarily across the entire formulation space. 
Three kinds of summaries are derived from the same profile.
\emph{Mean accessibility}
\refstepcounter{equation}\label{eq:mean-accessibility}%
$\mu=\mathbb{E}_{q\sim B}[A_q]$~(\theequation)
is the success probability for a random inquiry and formulation.
We take inquiries uniformly over the benchmark and write $q\sim B$ for brevity.
The \emph{reliability curve}
\begin{equation}
R(\tau)=\Pr_{q\sim B}[A_q\ge\tau]
\label{eq:reliability-threshold}
\end{equation}
gives the share of inquiries meeting an accessibility requirement $\tau$.
The area under the reliability curve equals the mean accessibility $\mu$, since an inquiry with
accessibility $a$ counts toward the curve for every $\tau\le a$
(formal proof in Appendix~\ref{app:proof-area}).
Its left end ($\tau\to0^+$) is the latent share $L=P(A>0)$, and
its right end is the reliable share $R=P(A=1)$. Finally, when a system faces $k$
independently sampled formulations, the probability that at least one formulation succeeds is
\begin{equation}
\mathrm{Access}@k = \mathbb{E}_q\bigl[1 - (1 - A_q)^k\bigr],
\end{equation}
and the probability that all $k$ formulations succeed would be
\begin{equation}
\mathrm{Reliable}@k = \mathbb{E}_q\bigl[A_q^k\bigr].
\end{equation} 
At $k=1$, both quantities reduce to mean accessibility $\mu$.
As $k\to\infty$, $\mathrm{Access}@k$ approaches the latent share
$L=P(A_q>0)$, since any inquiry with nonzero accessibility eventually succeeds
at least once. In contrast, $\mathrm{Reliable}@k$ approaches the reliable share
$R=P(A_q=1)$, since only inquiries with accessibility one can succeed on every
attempt indefinitely. Thus, the same accessibility profile captures deployments
that value repeated opportunity to succeed and those that require repeated
consistency.

An important point to note is that equal areas need not mean equal profiles. Two models
can have the same mean accessibility $\mu$, and hence the same expected success
on a random formulation, while distributing it very differently. A model whose
reliability curve starts high but falls quickly makes the desired behavior
reachable on many inquiries, but on few of them under every formulation; a model
whose curve starts lower but stays flat reaches fewer inquiries, yet what it
reaches holds up across formulations. $\mathrm{Access}@k$ rewards the first
shape and $\mathrm{Reliable}@k$ the second, and at $k=2$ the trade-off is exact:
$\mathrm{Access}@2+\mathrm{Reliable}@2=2\mu$. The shape of the curve decides
which model is preferred:

\begin{proposition}
\label{prop:single-crossing}
If two profiles $A$ and $A'$ have equal mean and their reliability curves cross
once, with $R(\tau)\le R'(\tau)$ below the crossing and $R(\tau)\ge R'(\tau)$
above it, then $\mathrm{Reliable}@k(A)\ge\mathrm{Reliable}@k(A')$ and
$\mathrm{Access}@k(A)\le\mathrm{Access}@k(A')$ for every $k\ge1$.
\end{proposition}

The proof is a standard convex-order argument (Appendix~\ref{app:proof-single-crossing}).
We estimate both quantities per inquiry without bias for $k\le m$, as
$\binom{c}{k}/\binom{m}{k}$ and $1-\binom{m-c}{k}/\binom{m}{k}$ for $c$ successes
among $m$ formulations.

\section{Experimental Setup}
\label{sec:experimental-setup}

\paragraph{Benchmarks and models.}
\label{sec:benchmarks-models}
We evaluate six benchmarks spanning factual question answering
(SimpleQA~\citep{wei2024simpleqa}, TruthfulQA~\citep{lin2022truthfulqa}),
mathematical reasoning (GSM8K~\citep{cobbe2021gsm8k},
MATH-500~\citep{hendrycks2021math,lightman2024verify}), knowledge
(MMLU~\citep{hendrycks2021mmlu}), and tool calling (BFCL~\citep{patil2025bfcl}),
with 18--21 non-thinking models per benchmark from the
Qwen3~\citep{yang2025qwen3}, Qwen2.5~\citep{qwen2024qwen25},
Gemma-3~\citep{gemma2025gemma3}, Llama~\citep{grattafiori2024llama3},
Mistral~\citep{jiang2023mistral}, Phi~\citep{li2023phi15}, and
OpenAI~\citep{openai2022chatgpt,openai2024gpt4o,openai2025gpt41,openai2026gpt54mini}
families, plus Qwen3 native-thinking variants (Table~\ref{tab:app-models} in
Appendix~\ref{app:models}).

\paragraph{Formulations.}
\label{sec:formulation-generation}
We generate candidate formulations of each inquiry with \texttt{GPT-4.1-mini}
using controlled structural and stylistic variation, remove near-verbatim
candidates, and retain only valid formulations that LLM judges verify as semantically
equivalent (Appendix~\ref{app:validation}). Human annotation of 1,072 pairs on
SimpleQA, TruthfulQA, and GSM8K agrees with the verification pipeline on
96.9--98.0\% of pairs (Table~\ref{tab:semantic-validation}). For each benchmark,
every model is evaluated on the same inquiries and the same formulations, and we
retain inquiries with at least five valid formulations: between 106 (MATH-500)
and 902 (GSM8K) inquiries per benchmark, with a median of 7--10 formulations each
(Table~\ref{tab:app-models}).

\paragraph{Scoring.}
\label{sec:evaluation-scoring}
Each model answers every retained formulation under the fixed configuration,
with greedy decoding unless stated otherwise. Responses are scored with
benchmark-specific binary scorers (symbolic answer verification for math,
answer extraction for MMLU, multi-judge grading for open-ended QA, and
structured checking of tool calls), and
Eq.~\ref{eq:accessibility-estimate} aggregates them per inquiry.
\section{Results}
\label{sec:results}

\subsection{RQ1: What Does the Accessibility Profile Reveal That Accuracy Hides?}
\label{sec:rq1}

Benchmark accuracy is used to decide whether a model can perform a task and which
of two models is better. Both uses assume that one wording of each inquiry is
representative of how the model responds to it. We test this assumption by
comparing each canonical outcome $Y_{\mathrm{orig}}(q)$ with the estimated
accessibility $\hat A_q$, and then ask whether even the mean of the profile
suffices to compare models.

\begin{figure*}[t]
    \centering
    \includegraphics[width=\textwidth]{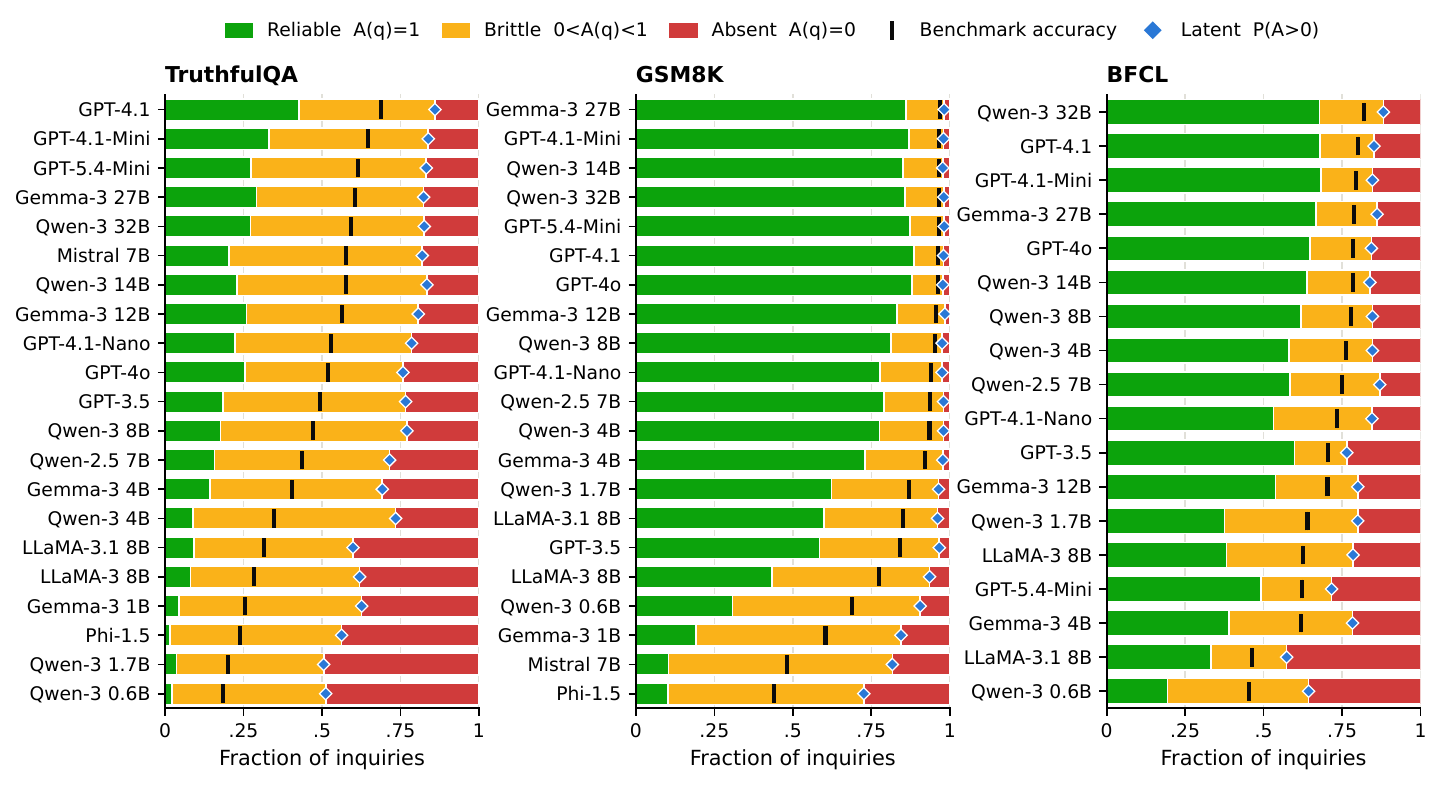}
    \caption{
    \textbf{Benchmark accuracy is one point inside a capability envelope.}
    Inquiries per model are reliable ($\hat A_q=1$), brittle ($0<\hat A_q<1$), or
    absent ($\hat A_q=0$). Black ticks mark benchmark accuracy and blue diamonds
    the latent share $P(\hat A_q>0)$; models are ordered by accuracy. All
    non-thinking models on a shared inquiry set per benchmark; SimpleQA,
    MATH-500, and MMLU in Figure~\ref{fig:app-envelope-all}.
    }
    \label{fig:capability-envelope}
\end{figure*}

\paragraph{Accuracy is one point inside a wide envelope.}
Because the canonical formulation is among the evaluated formulations, every
reliable inquiry (solved under every formulation, $\hat A_q=1$) is a canonical
success and every canonical success is latent (solved under at least one,
$\hat A_q>0$), so $R\le\mathrm{Acc}\le L$. Accuracy is one point inside this
envelope. Its width, $L-R$, is the brittle share ($0<\hat A_q<1$): inquiries
whose outcome depends on the wording. The envelope is wide (Figure~\ref{fig:capability-envelope}): the median
brittle share is 55\% on TruthfulQA, 37\% on MMLU, and 20--24\% on GSM8K,
MATH-500, and BFCL, with SimpleQA, where most inquiries are absent 
($\hat A_q=0$) for every model, the low-capability exception at 6\% (Table~\ref{tab:rq1-summary} in
Appendix~\ref{app:rq1-full}).

\begin{figure*}[t]
    \centering
    \includegraphics[width=\textwidth]{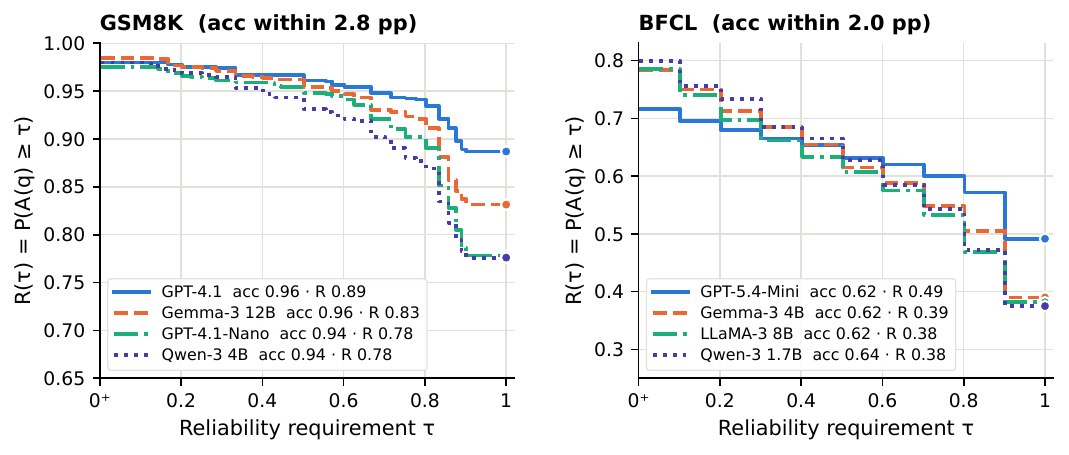}
    \caption{
    \textbf{Similar accuracy can conceal different reliability profiles.}
    Reliability curves $R(\tau)=P(A_q\geq\tau)$ for near-tied models on GSM8K
    (left; accuracy within 2.8\,pp) and BFCL (right; within 2.0\,pp). The left
    end is the latent share, the right end (dot) the reliable share.
    }
    \label{fig:rq1-reliability}
\end{figure*}

\paragraph{Even mean accessibility is not enough: the shape of the profile decides.}
Averaging over formulations does not remove the problem, because models with the
same mean accessibility can distribute it very differently.
Figure~\ref{fig:rq1-reliability} shows reliability curves
(Eq.~\ref{eq:reliability-threshold}) for 
four near-tied models (in terms of accuracy) on GSM8K and four
on BFCL. On GSM8K, the four models are within 2.8 points of accuracy and have
nearly identical latent shares, yet their reliable shares differ by up to 11.1
points (95\% CI $[8.6,13.7]$), a gap that appears only as the requirement
approaches $\tau=1$. On BFCL, the curves cross: GPT-5.4-Mini has a lower latent
but a higher reliable share than Qwen-3 1.7B at nearly equal mean accessibility.
By Proposition~\ref{prop:single-crossing}, GPT-5.4-Mini must therefore rank higher
under every $\mathrm{Reliable}@k$ and lower under every $\mathrm{Access}@k$; at
$k=5$ it leads by 9.6 points ($[5.5,14.0]$) and trails by 6.8 points
($[-10.8,-2.8]$). Which model is better depends on whether the deployment needs
one success or success every time, and neither accuracy nor mean accessibility
can tell.

\paragraph{Such reversals are common.}
Among the 118 model pairs whose mean accessibility differs by at most 2 points
(TruthfulQA, GSM8K, MATH-500, MMLU, BFCL), $\mathrm{Access}@5$ and
$\mathrm{Reliable}@5$ rank the two models in opposite orders for about half
(53\%). For 11 pairs both gaps exceed 2 points, 8 of them significantly in both
directions, and 10 of these 11 have reliability curves that cross exactly once,
as Proposition~\ref{prop:single-crossing} describes. The largest case is
LLaMA-3 8B versus LLaMA-3.1 8B on MMLU: with mean accessibility within 0.1
points, the former is 10.0 points higher in $\mathrm{Reliable}@5$
($[6.3,13.8]$) and 9.3 points lower in $\mathrm{Access}@5$ ($[-12.8,-5.8]$).
Accuracy performs no better. Across all models it ranks similarly to the reliable
share (Kendall's $\tau$ of 0.83--0.93 per benchmark), but models are compared at
a resolution of a few points, and there the two come apart. Consider the 164
model pairs whose accuracies are within 3 points of each other. In 143 of them
(87\%), the two reliable shares differ by more than the two accuracies. In 40
(24\%), the more accurate model has the \emph{smaller} reliable share,
significantly so in 16 (Figure~\ref{fig:app-near-tied} in
Appendix~\ref{app:rq1-near-tied}).

\paragraph{The canonical label can even point the wrong way.}
For some inquiries the canonical outcome lies on the minority side of the model's
behavior: the benchmark records a success although most equivalent formulations
fail, or a failure although most succeed. Such majority reversals affect a median
of 15.3\% of TruthfulQA and 11.3\% of MMLU inquiries, and up to 23.6\% for an
individual model, in both directions  (Table~\ref{tab:rq1-summary},
Figures~\ref{fig:app-reversals}--\ref{fig:app-reversals-cond}).
For these inquiries the label gives
the opposite impression of how the model typically behaves, and any conclusion
drawn from them individually, such as an error analysis, inherits this error.

\subsection{RQ2: Is Formulation Variability More Than Decoding Noise?}
\label{sec:rq2}

\begin{figure*}[t]
    \centering
    \includegraphics[width=\textwidth]{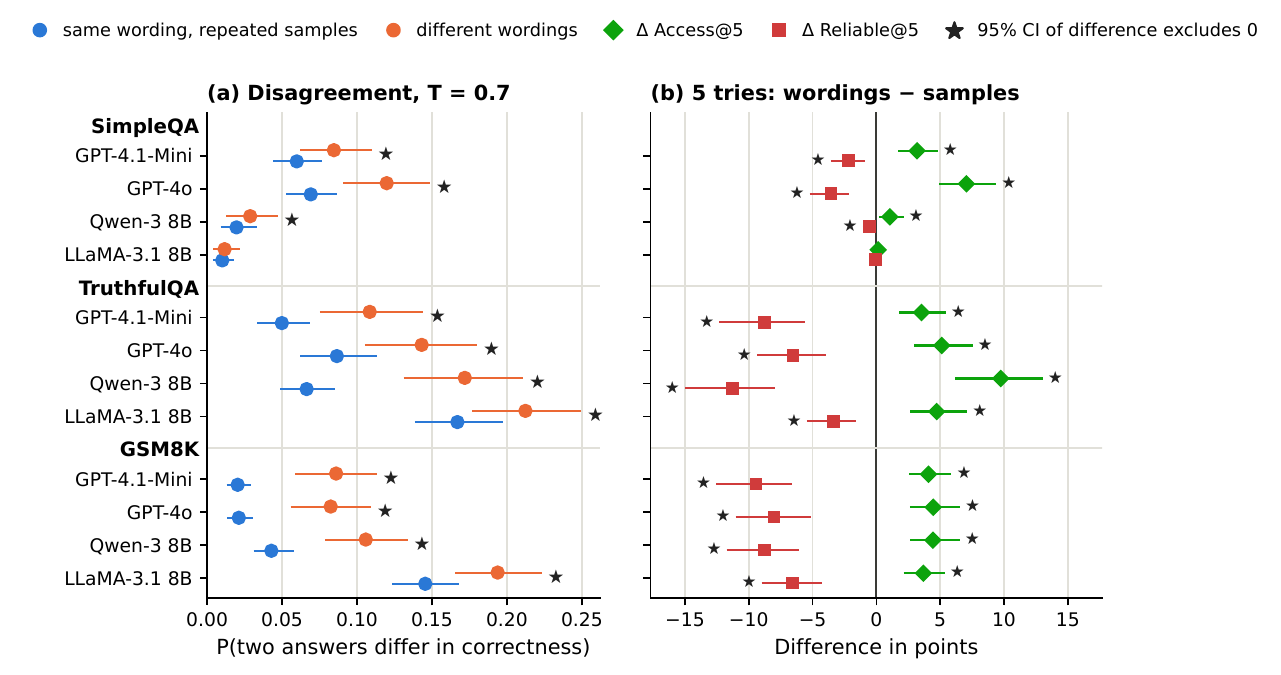}
    \caption{
    \textbf{Changing the wording changes behavior more than resampling it.}
    Rows are model--benchmark settings; $T=0.7$. (a)~Probability that two
    answers differ in correctness, for two samples of the same wording (blue) and
    for two different equivalent wordings (orange). (b)~With five attempts:
    $\mathrm{Access}@5$ and $\mathrm{Reliable}@5$ over five different wordings
    minus the same over five samples of one wording. Only paraphrases count as
    different wordings. Bars: 95\% bootstrap CIs over inquiries. $\star$: the 95\%
    CI of the difference excludes zero.
    }
    \label{fig:rq2}
\end{figure*}

The structure in RQ1 is meaningful only if changing the wording changes behavior
more than simply sampling again does. In a controlled sub-study, four models
(GPT-4.1-Mini, GPT-4o, Qwen-3 8B, LLaMA-3.1 8B) answer 150 SimpleQA, 102
TruthfulQA, and 150 GSM8K inquiries under seven equivalent formulations each,
with five samples per formulation at temperature $0.7$. This lets us compare,
under identical sampling, answers to the same wording with answers to different
wordings (Figure~\ref{fig:rq2}).

\paragraph{Different wordings disagree more than repeated samples.}
Two answers to different formulations differ in correctness more often than two
samples of the same formulation (both at $T=0.7$) in 11 of 12 settings (95\% CI
of the difference excludes zero; Figure~\ref{fig:rq2}); on TruthfulQA, for example, Qwen-3 8B disagrees with itself in
6.6\% of repeated samples but in 17.2\% of answers to different wordings. The
exception is LLaMA-3.1 8B on SimpleQA, which answers about 1\% of inquiries
correctly. The same holds under the greedy decoding used in our main
experiments: repeated greedy runs of one wording differ in correctness in
0.2--4.9\% of cases, reflecting nondeterminism in the inference backends (the generated text itself changes), whereas
different wordings differ in up to 16.4\% (Figure~\ref{fig:app-rq2-greedy} and
Table~\ref{tab:app-rerun} in Appendix~\ref{app:rq2-full}).

\paragraph{Varying the wording is both a more generous and a stricter test.}
With a budget of five attempts, spreading them over five wordings rather than
five samples of one wording raises $\mathrm{Access}@5$ by 1.1 to 9.7 points and
lowers $\mathrm{Reliable}@5$ by 0.5 to 11.3 points, both significantly in 11 of
12 settings (Figure~\ref{fig:rq2}b, also other budgets in Figure~\ref{fig:app-rq2-k}). Repeatedly sampling one prompt, as pass@$k$
and pass$^k$ usually do, therefore underestimates the capability a model can
reach and overestimates how reliably it delivers it to users who rephrase.
Accessibility therefore measures structure induced by the choice of formulation,
not a repackaging of decoding stochasticity.

\subsection{RQ3: What Do Accessibility Profiles Reveal About Model Changes?}
\label{sec:rq3}

Whether a model change helps is usually decided by its change in benchmark
accuracy. We compare, for every inquiry, the canonical label and the
accessibility before and after 119 model changes, each on one benchmark: 48
scaling steps (Qwen3, Gemma-3, GPT-4.1), 30 switches to native thinking (Qwen3),
21 switches to chain-of-thought prompting, and 20 prompt optimizations (GEPA,
MIPROv2).\footnote{Scaling and native thinking use the RQ1 inquiry sets; prompted
CoT (TruthfulQA) and prompt optimization (GSM8K, TruthfulQA) are sub-studies on
separately generated formulation sets.} 

\begin{figure*}[t]
    \centering
    \includegraphics[width=\textwidth]{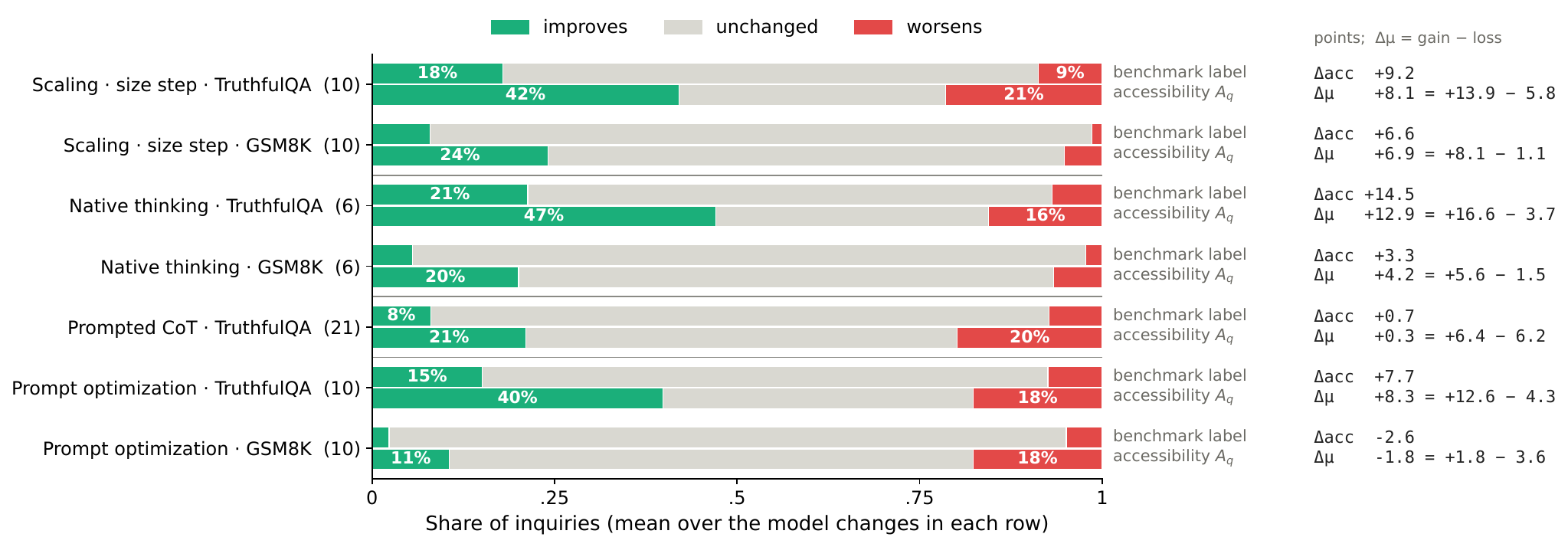}
    \caption{
    \textbf{Canonical labels reveal only part of what changes under an
    intervention.} Per row, the upper bar shows the share of inquiries whose
    canonical outcome improves, stays the same, or worsens; the lower bar the
    share whose accessibility $A_q$ increases, stays the same, or decreases
    (means over the model changes in the row, count in parentheses). Rows are the
    benchmarks on which every intervention was run; all benchmarks in
    Appendix~\ref{app:rq3-full}. Right: change in accuracy and in mean
    accessibility, $\Delta\mu$ = mean gain $-$ mean loss.
    }
    \label{fig:rq3-two-views}
\end{figure*}

\paragraph{Model changes redistribute capability rather than simply adding or removing it.}
Figure~\ref{fig:rq3-two-views} compares how often canonical labels change with how often inquiry accessibility changes. Accessibility changes are much more common than label changes, and inquiries can become either more or less accessible. For example, on GSM8K, scaling changes the accessibility of 29\% of inquiries on average (24\% improve and 5\% worsen), and native thinking changes 27\% (20\% improve and 7\% worsen) (Figure~\ref{fig:rq3-two-views}).
The results for native thinking on MMLU show that a small change in overall accuracy can hide much larger changes across individual inquiries (Figure~\ref{fig:app-rq3-two-views} in Appendix~\ref{app:rq3-full}). Averaged over the six Qwen3 sizes, accuracy drops by only 3.0 points, but 25\% of inquiries become more accessible and 32\% become less accessible. The reliable share also drops by 10.9 points.

\begin{figure*}[t]
    \centering
    \includegraphics[width=\textwidth]{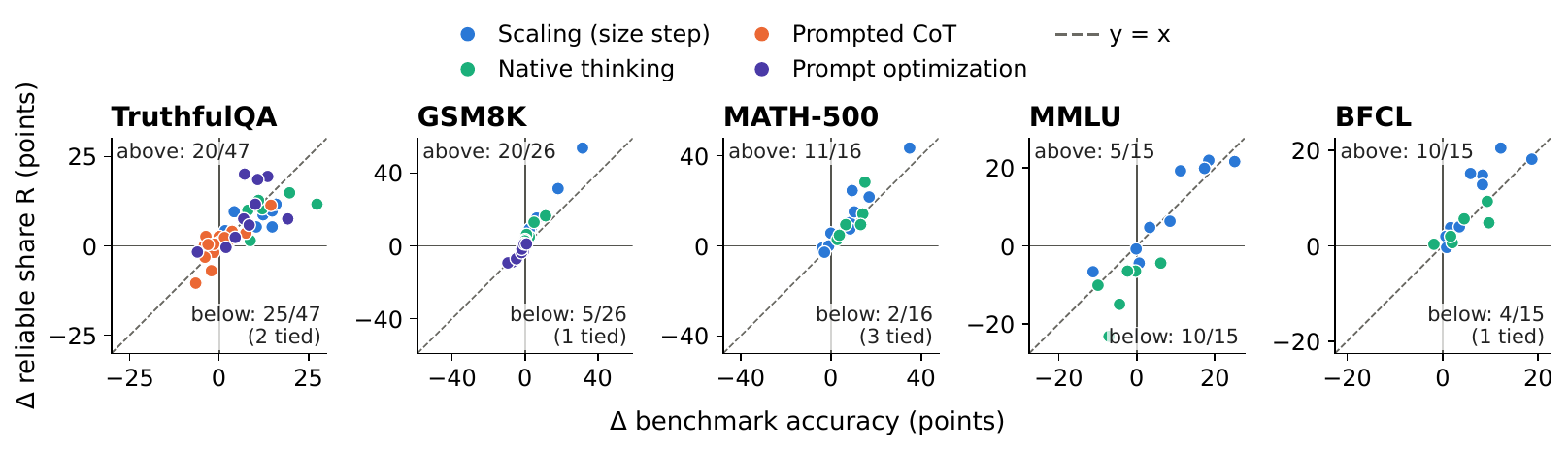}
    \caption{
    \textbf{The same change in accuracy can correspond to very different changes
    in reliable capability.} Each point is one model change; x: change in
    benchmark accuracy, y: change in reliable share $R=P(\hat A_q=1)$. Points
    above the dashed line gain reliable capability faster than accuracy (or lose
    it more slowly). Counts give the changes on each side of the diagonal and
    exact ties.
    }
    \label{fig:rq3-dacc-dr}
\end{figure*}

\paragraph{Accuracy does not reveal what kind of capability changed.}
The redistribution also changes reliability in ways accuracy does not track
(Figure~\ref{fig:rq3-dacc-dr}). On GSM8K, the reliable share grows faster than
accuracy for all ten scaling steps and all six native-thinking comparisons, so
moderate accuracy gains correspond to larger gains in inquiries answered
correctly under every wording. On MMLU, native thinking does the opposite for all
six Qwen3 sizes: Qwen3-8B gains 6.2 points of accuracy but loses 4.4 points of
reliable share, and Qwen3-32B loses 7.1 points of accuracy but 23.2 of reliable
share. More broadly, most changes lie above the diagonal on GSM8K, MATH-500, and
BFCL (20 of 26, 11 of 16, and 10 of 15), whereas 10 of 15 lie below it on MMLU.
The same accuracy change can therefore mean that a model became more reliable,
or only that its performance on the canonical wording shifted.

\begin{figure*}[t]
    \centering
    \includegraphics[width=\textwidth]{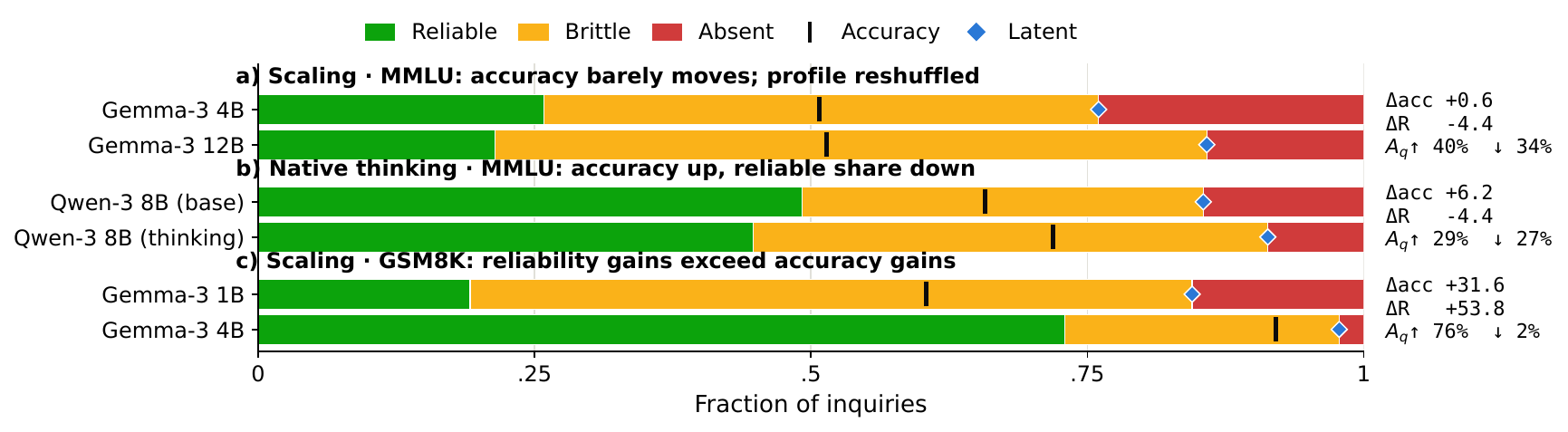}
    \caption{
    \textbf{Accessibility profiles before and after a model change.} Each pair
    of bars shows the model before (upper) and after (lower) the change, with the
    same encoding as Figure~\ref{fig:capability-envelope}. Right: change in
    accuracy and in reliable share (points), and the share of inquiries whose
    accessibility increases or decreases.
    }
    \label{fig:rq3-paired}
\end{figure*}

Figure~\ref{fig:rq3-paired} shows such redistributions directly, in the
encoding of Figure~\ref{fig:capability-envelope}: a profile reshuffled at
almost unchanged accuracy, accuracy rising while the reliable share falls, and
reliability gains far exceeding accuracy gains. Paired profiles for all 119 model
changes are in Appendix~\ref{app:rq3-paired}.

\section{Discussion}
\label{sec:discussion}

A formulation is an observation; the inquiry is the object being evaluated; its accessibility $A_q$ is the quantity to estimate; and the questions practitioners ask correspond to different summaries of the same accessibility profile, of which single-formulation accuracy is one noisy observation.

\paragraph{What evaluations should report.}
Our results suggest treating correctness as a distribution over semantically
equivalent formulations rather than as a single outcome per item. Reporting the
reliable share $R$ and the latent share $L$ alongside accuracy exposes the width
of the envelope that accuracy collapses, and the reliability curve $R(\tau)$
shows where between these endpoints models differ. 
This need not multiply the evaluation cost: in our benchmarks, evaluating two formulations on half of the inquiries, at the cost of one conventional run, recovered mean accessibility about as precisely as accuracy on every benchmark except the small MATH-500, and the two-formulation latent and reliable shares for every model on the four benchmarks with 600 or more inquiries (Appendix~\ref{app:budget}).

\paragraph{Why aggregate rankings look reassuring.}
Rewording leaves coarse model rankings largely intact
\citep{robustness2025benchmark}, and so do our data: accuracy and mean
accessibility order models almost identically (Kendall's $\tau$ of 0.90--0.99),
whereas the reliable and latent shares agree only at $\tau$ of 0.56--0.85. Averaging
preserves the coarse ordering but discards the shape of the profile, which decides
whether capability is broad but brittle or narrow but dependable, and whether
near-tied models reverse between deployment regimes (RQ1). Ranking stability is
not evidence of behavioral equivalence.


\paragraph{Deployment regimes and pipeline design.}
Accessibility profiles provide a practical diagnosis of how a model's capability
is distributed: whether success is broadly reachable but brittle, or narrower but
dependable. This distinction matters because different deployment regimes value
these behaviors differently. When failures are cheap to detect and retry (e.g.,
a verifier, test suite, or human checks the output), the relevant quantity is
$\mathrm{Access}@k$; when failures may go undetected or outputs feed later
components automatically, it is $\mathrm{Reliable}@k$.
Proposition~\ref{prop:single-crossing} formalizes when two models with similar
average performance are necessarily preferred in opposite regimes.

The profile also informs the retry budget: an inquiry with accessibility $A_q$
requires $1/A_q$ attempts on average to obtain a success.
The expected number of attempts to solve an arbitrary query sampled from a benchmark B therefore depends on the whole profile, not only on the mean accessibility (Eq.~\ref{eq:mean-accessibility}), since $\mathbb{E}_{q\sim B}[1/A_q]\neq 1/\mathbb{E}_{q\sim B}[A_q]$.
On BFCL, Qwen-3 1.7B
and GPT-5.4-Mini have nearly equal mean accessibility, yet Qwen-3 1.7B requires
1.92 attempts per solvable inquiry versus 1.50 for GPT-5.4-Mini, about 28\% more
calls, in exchange for reaching more inquiries. Thus, in agentic systems built
from many model calls, accessibility profiles can inform not only which model to
use, but also where retries or verification are worth the cost and how the
available token/compute budget should be allocated across pipeline components.

\paragraph{Evaluating model changes.}
A change in accuracy is the net result of gains and losses across inquiries
(RQ3). For model updates, we recommend reporting the gains and losses behind
$\Delta\mu$ and tracking per-inquiry regressions, extending negative-flip counting \citep{yan2021pct,echterhoff2024muscle}
from single outputs to accessibility.

\paragraph{Limitations.}
Accessibility is estimated from a finite set of formulations that approximates,
but does not sample, the distribution of real user phrasings; weighting
formulations by observed frequency is a natural extension. Endpoint labels (reliable, absent) are therefore empirical: with more
formulations, an inquiry can only move from reliable or absent to brittle, so
the reported brittle shares are lower bounds with respect to the formulation
distribution; conversely, inference-backend nondeterminism adds some variation
that is not caused by the wording, which we quantify for GSM8K (RQ2, RQ3) but not
for every benchmark. RQ2 and the prompted-CoT and prompt-optimization analyses are smaller sub-studies. Finally, we use
binary task scoring and English benchmarks; extending the framework to
graded scores, other languages, and other modalities is future work.
\section{Conclusion}
\label{sec:conclusion}


Prior work established that language models are sensitive to how a request is
phrased. We treat this variation as information about the evaluated capability:
each inquiry has an accessibility, and a model has an accessibility profile. The
profile reveals what single-formulation accuracy hides, from reliability
differences between near-tied models to redistributions of capability under model
changes, and reporting it gives a more faithful basis for comparing, selecting,
and updating models.

\section*{Code and data availability}
The \texttt{access-eval} toolkit, the validated formulation sets, and the code and data that reproduce every
figure, table, and number in this paper are available at \url{https://github.com/kazemf78/access-eval}. The full
model answers and judge grades are available at \url{https://github.com/kazemf78/access-eval-data}.

\section*{Acknowledgments}
This work was supported in part by NSF CAREER Award 1942230, the ONR PECASE Award N00014-25-1-2378, ARO Early Career Program Award 310902-00001, Army Grant W911NF-21-2-0076, NSF Award CCF-2212458, NSF Award 2229885 (NSF Institute for Trustworthy AI in Law and Society, TRAILS), MURI Grant 14262683, DARPA AIQ Grant HR00112590066, an award from Google Research, and an award from Meta 314593-00001.

\bibliography{iclr2027_conference}
\bibliographystyle{iclr2027_conference}

\appendix
\section{Proofs for Section~\ref{sec:accessibility-profiles}}

\subsection{Area Under the Reliability Curve}
\label{app:proof-area}

For every inquiry, $A_q=\int_0^1\mathbf{1}\{A_q\ge\tau\}\,d\tau$, because the
indicator equals $1$ for $\tau\in[0,A_q]$ and $0$ otherwise. Taking the
expectation over inquiries and exchanging expectation and integral (the
integrand is bounded),
\[
\mu=\mathbb{E}_q[A_q]
=\int_0^1\mathbb{E}_q\bigl[\mathbf{1}\{A_q\ge\tau\}\bigr]\,d\tau
=\int_0^1\Pr_q\bigl[A_q\ge\tau\bigr]\,d\tau
=\int_0^1 R(\tau)\,d\tau .
\]
The area under the reliability curve is therefore exactly the mean
accessibility. \qed

\subsection{Proof of Proposition~\ref{prop:single-crossing}}
\label{app:proof-single-crossing}

Write $\mathbb{E}[\cdot]$ for expectation over inquiries, $R(t)=P(A\ge t)$, and
$R'(t)=P(A'\ge t)$. For any twice-differentiable $\varphi$ on $[0,1]$, Taylor's
theorem with integral remainder gives, for $a\in[0,1]$,
\[
\varphi(a)=\varphi(0)+\varphi'(0)\,a+\int_0^1\varphi''(t)\,(a-t)_+\,dt ,
\]
and $\mathbb{E}[(A-t)_+]=\int_t^1 R(s)\,ds$. Taking expectations and
subtracting the same identity for $A'$, the linear terms cancel because
$\mathbb{E}[A]=\mathbb{E}[A']$, so
\[
\mathbb{E}[\varphi(A)]-\mathbb{E}[\varphi(A')]
=\int_0^1\varphi''(t)\,H(t)\,dt,
\qquad
H(t)=\int_t^1\bigl(R(s)-R'(s)\bigr)\,ds .
\]
Here $H(1)=0$, and $H(0)=\mathbb{E}[A]-\mathbb{E}[A']=0$ because
$\int_0^1 R(s)\,ds=\mathbb{E}[A]$. Since $H'(t)=R'(t)-R(t)$, the single-crossing
assumption makes $H$ non-decreasing on $[0,\tau^\ast]$ and non-increasing on
$[\tau^\ast,1]$. Together with $H(0)=H(1)=0$, this gives $H\ge0$ on $[0,1]$.
(This is the Karlin--Novikoff cut criterion: $A$ dominates $A'$ in the convex
order.)

For $\varphi(a)=a^k$, $\varphi''(t)=k(k-1)t^{k-2}\ge0$, hence
$\mathrm{Reliable}@k(A)-\mathrm{Reliable}@k(A')\ge0$. For
$\varphi(a)=1-(1-a)^k$, $\varphi''(t)=-k(k-1)(1-t)^{k-2}\le0$, hence
$\mathrm{Access}@k(A)-\mathrm{Access}@k(A')\le0$. At $k=1$ both differences are
$\mathbb{E}[A]-\mathbb{E}[A']=0$. \qed

\paragraph{Remarks.}
(i) The identity $\mathrm{Access}@2+\mathrm{Reliable}@2=2\mu$ follows from
$1-(1-a)^2+a^2=2a$. (ii) The proposition concerns population profiles. With
finitely many formulations, we apply it to the empirical profiles and report
the unbiased per-inquiry estimators with bootstrap confidence intervals.
(iii) The effective-breadth/repeatability decomposition of
Appendix~\ref{app:breadth-repeatability} is the $k=2$ case of this view:
$D=\mathrm{Reliable}@2/\mu$.

\section{Formulation Construction and Validation}

\label{app:validation}

This section provides implementation and validation details for the formulation-construction procedure described in Section~\ref{sec:formulation-generation}. Our goal is to construct formulation sets that satisfy two requirements: they should exhibit meaningful linguistic variation while preserving the same underlying inquiry and desired behavior.

\subsection{Controlled Formulation Generation}
\label{app:paraphrase-generation}

For each original inquiry, we generate candidate formulations using \texttt{GPT-4.1-mini}. Rather than relying on unconstrained paraphrasing, generation is guided by lightweight, non-adversarial cues designed to vary surface form while preserving meaning, assumptions, intent, and required information.

We use two complementary types of generation cues.

\paragraph{Structural variation.}
Structural cues encourage changes in wording and syntax without introducing a specific style or tone. We randomly sample from a pool of transformations including reordering clauses, changing sentence openers, varying syntactic structure, substituting safe synonyms, compressing or expanding phrasing, switching between active and passive voice, and changing between statement and question form where appropriate.

\paragraph{Stylistic variation.}
We additionally generate formulations conditioned on stylistic attributes including \emph{formal}, \emph{casual}, \emph{polite}, \emph{confident}, \emph{neutral}, \emph{supportive}, and \emph{inquisitive}. These cues are intended to capture natural variation in how users may express the same request rather than adversarial or performance-targeted perturbations.

\begin{figure}[H]
\centering
\begin{tcolorbox}[
enhanced,
colframe=black!50,
colback=white,
width=\linewidth,
arc=2mm,
boxrule=0.5pt,
left=10pt,
right=10pt,
top=8pt,
bottom=8pt,
title=\textbf{Prompt for Structural Paraphrasing},
fonttitle=\bfseries
]
You are a careful, concise rewriter. \\
Paraphrase the user's question below without changing its meaning, intent, or sentiment.
\begin{itemize}
\item Use different wording and structure.
\item Keep the output natural and fluent.
\item Output \textbf{only} the final paraphrase as a single line.
\end{itemize}
Diversity tip: [structural tip] \\
Question: [original question]
\end{tcolorbox}
\label{fig:plain_paraphrase_prompt}
\end{figure}

\begin{figure}[H]
\centering
\begin{tcolorbox}[
enhanced,
colframe=black!50,
colback=white,
width=\linewidth,
arc=2mm,
boxrule=0.5pt,
left=10pt,
right=10pt,
top=8pt,
bottom=8pt,
title=\textbf{Prompt for Stylistic Paraphrasing},
fonttitle=\bfseries
]
You are a careful, concise rewriter. \\
Rewrite the user's question with the requested tone or attribute, but do \emph{not} change its meaning, intent, assumptions, or sentiment.
\begin{itemize}
\item Use different wording and structure.
\item Keep the output natural and fluent.
\item Output \textbf{only} the final paraphrase as a single line.
\end{itemize}
Tone/attribute: [tone] --- [description] \\
Question: [original question]
\end{tcolorbox}
\label{fig:toned_paraphrase_prompt}
\end{figure}

\subsection{Diversity Filtering}
\label{app:diversity-filtering}

Because inquiry accessibility is estimated across alternative formulations, near-verbatim rewrites provide little additional information about the formulation space. We therefore filter candidates based on lexical overlap with the original formulation.

We define lexical overlap as the proportion of shared word tokens between the candidate and original formulation, normalized by the length of the shorter text. Candidates with lexical overlap above $0.8$ are discarded. This removes near-duplicate rewrites while retaining formulations with meaningful lexical and structural variation.


\subsection{Semantic Verification}
\label{app:semantic-verification}

Diversity alone is insufficient: candidate formulations must also preserve the same underlying inquiry and desired behavior. We therefore apply semantic verification after diversity filtering.

\paragraph{Independent LLM judges.}
Candidate formulations are evaluated for semantic equivalence using \texttt{GPT-5} and \texttt{GPT-5.4-mini}. Candidates are retained only when the required judges agree that they express the same inquiry as the original formulation. Because formulation generation uses \texttt{GPT-4.1-mini}, generation and semantic verification are performed by different models.

The semantic-equivalence prompt is:

\begin{figure}[H]
\centering
\begin{tcolorbox}[
enhanced,
colframe=black!50,
colback=white,
width=\linewidth,
arc=2mm,
boxrule=0.5pt,
left=10pt,
right=10pt,
top=8pt,
bottom=8pt,
title=\textbf{Prompt for Semantic-Equivalence Verification},
fonttitle=\bfseries
]
You are a precise evaluator of meaning. \
Determine if the candidate paraphrase expresses the SAME intent and meaning as the original question, without adding or removing constraints or changing sentiment. \
Could they be used interchangeably in the same context? If they are questions, do they ask the same thing, carry the same assumptions, and admit the same answer? \

Original: [original] \
Candidate: [candidate] \

Respond with EXACTLY one token: YES or NO.
\end{tcolorbox}
\label{fig:judge_equivalence_prompt}
\end{figure}

\paragraph{Bidirectional NLI filtering.}
We additionally require bidirectional entailment using \texttt{facebook/bart-large-mnli}. For an original formulation $Q^0$ and candidate $Q^i$, we compute entailment in both directions and retain the candidate only when both scores exceed $0.75$. This additional constraint helps reject candidates that omit information, introduce new constraints, or otherwise alter the inquiry.
The final formulation sets therefore satisfy both diversity and semantic-preservation criteria. 

\subsection{Human Validation of Semantic Equivalence}
\label{app:human-validation}

We independently evaluate whether retained formulations preserve the same underlying inquiry through human annotation. Across 1,072 formulation pairs distributed among five annotators, each pair receives a binary YES/NO judgment indicating whether the two formulations ``ask the exact same thing,'' i.e., whether they can be used interchangeably without changing the desired answer or behavior.

Agreement between human judgments and the automatic semantic-verification pipeline is consistently high, ranging from 96.9\% to 98.0\% across benchmarks.

\begin{table}[H]
\centering
\caption{Human validation of formulation equivalence. We report the number of annotated formulation pairs and agreement between human judgments and the automatic semantic-verification pipeline.}
\label{tab:semantic-validation}
\begin{tabular}{lcc}
\toprule
\textbf{Benchmark} & \textbf{Pairs} & \textbf{Agreement} \\
\midrule
SimpleQA   & 460 & 96.9\% \\
TruthfulQA & 232 & 98.0\% \\
GSM8K      & 380 & 97.7\% \\
\bottomrule
\end{tabular}
\end{table}

The reported agreement measures \emph{human--pipeline agreement}, not inter-annotator agreement; specifically, it is the fraction of annotated pairs for which the human judgment matches the final decision of the semantic-verification pipeline.

\subsection{Summary}

Our construction procedure targets two complementary requirements for inquiry-level evaluation. Diversity filtering ensures that sampled formulations contain meaningful variation in expression, while semantic verification ensures that this variation does not change the underlying inquiry or desired behavior. High agreement with human judgments provides additional evidence that the resulting formulation sets preserve semantic equivalence.

\section{Benchmarks and Models}
\label{app:models}

Table~\ref{tab:app-models} lists the non-thinking models evaluated on each
benchmark in RQ1 (all models on a benchmark share the same inquiries and
formulations). A formulation is valid for a benchmark if every model answered it
and every answer could be scored (for open-ended QA, the judges agree); we retain
inquiries with at least five valid formulations. Gemma-3 1B is not evaluated on MMLU and BFCL, and Mistral 7B and
Phi-1.5 are not evaluated on BFCL.
RQ3 additionally uses the native-thinking variants of the six Qwen3 models on
TruthfulQA, GSM8K, MATH-500, MMLU, and BFCL. The prompted-CoT sub-study uses the
same 21 models on TruthfulQA, the prompt-optimization sub-study uses Qwen-3 8B,
Gemma-3 4B, LLaMA-3.1 8B, GPT-4.1-Mini, and GPT-5.4-Mini on GSM8K and TruthfulQA,
and the RQ2 sub-study uses GPT-4.1-Mini, GPT-4o, Qwen-3 8B, and LLaMA-3.1 8B.

\begin{table}[h]
    \centering
    \small
    \caption{Models per benchmark in RQ1 (\checkmark = evaluated). Inquiries
    (median formulations per inquiry): SimpleQA 618 (7), TruthfulQA 471 (9),
    GSM8K 902 (7), MATH-500 106 (8), MMLU 634 (7), BFCL 600 (10).}
    \label{tab:app-models}

    \resizebox{\linewidth}{!}{%
    \begin{tabular}{lcccccc}
    \toprule
    Model & SimpleQA & TruthfulQA & GSM8K & MATH-500 & MMLU & BFCL \\
    \midrule
    Qwen-3 0.6B / 1.7B / 4B / 8B / 14B / 32B & \checkmark & \checkmark & \checkmark & \checkmark & \checkmark & \checkmark \\
    Qwen-2.5 7B & \checkmark & \checkmark & \checkmark & \checkmark & \checkmark & \checkmark \\
    Gemma-3 1B & \checkmark & \checkmark & \checkmark & \checkmark & -- & -- \\
    Gemma-3 4B / 12B / 27B & \checkmark & \checkmark & \checkmark & \checkmark & \checkmark & \checkmark \\
    LLaMA-3 8B, LLaMA-3.1 8B & \checkmark & \checkmark & \checkmark & \checkmark & \checkmark & \checkmark \\
    Mistral 7B, Phi-1.5 & \checkmark & \checkmark & \checkmark & \checkmark & \checkmark & -- \\
    GPT-3.5, GPT-4o & \checkmark & \checkmark & \checkmark & \checkmark & \checkmark & \checkmark \\
    GPT-4.1 / Mini / Nano, GPT-5.4-Mini & \checkmark & \checkmark & \checkmark & \checkmark & \checkmark & \checkmark \\
    \midrule
    Models & 21 & 21 & 21 & 21 & 20 & 18 \\
    \bottomrule
    \end{tabular}%
    }
\end{table}

\section{Additional RQ1 Results}
\label{app:rq1-full}

\begin{table}[h]
    \centering
    \small
    \begin{tabular}{lrr|ccc|cc}
\toprule
Benchmark & \#M & \#Inq & Acc & $R$ & $L$ & Brittle ($L-R$) & Majority reversal \\
\midrule
SimpleQA & 21 & 618 & 2.9 & 0.5 & 6.5 & 6.0 (max 27.5) & 2.1 (max 9.1) \\
TruthfulQA & 21 & 471 & 49.3 & 18.5 & 76.6 & 55.0 (max 64.5) & 15.3 (max 18.5) \\
GSM8K & 21 & 902 & 93.8 & 77.8 & 97.8 & 19.7 (max 71.3) & 3.4 (max 23.6) \\
MATH-500 & 21 & 106 & 84.0 & 68.9 & 93.4 & 23.6 (max 59.4) & 4.7 (max 20.8) \\
MMLU & 20 & 634 & 70.6 & 46.1 & 87.2 & 37.0 (max 67.8) & 11.3 (max 22.6) \\
BFCL & 18 & 600 & 74.2 & 58.2 & 84.1 & 23.5 (max 45.0) & 4.9 (max 9.0) \\
\bottomrule
\end{tabular}

    \caption{
    \textbf{Per-benchmark summary} (medians over models, \%). $R=P(\hat A=1)$ is
    the reliable share and $L=P(\hat A>0)$ the latent share; accuracy always lies
    in $[R,L]$, and the brittle share $L-R$ is the width of that interval.
    Majority reversal is the share of inquiries whose canonical outcome disagrees
    with the majority of formulations. Maxima over models in parentheses.
    }
    \label{tab:rq1-summary}
\end{table}

\begin{figure*}[h]
    \centering
    \includegraphics[width=\textwidth]{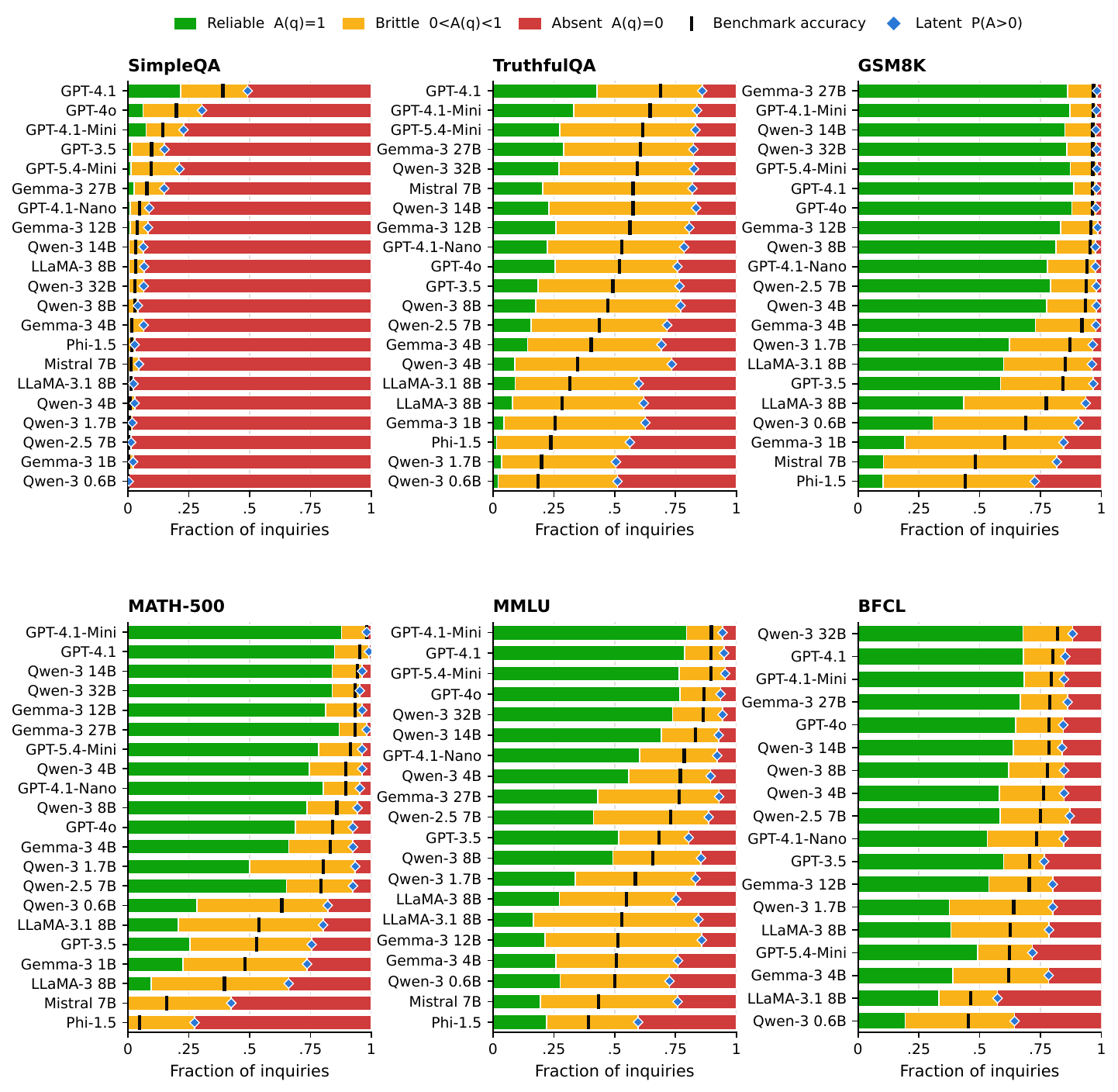}
    \caption{Capability envelopes for all six benchmarks, including SimpleQA and MATH-500 (as in Figure~\ref{fig:capability-envelope}).}
    \label{fig:app-envelope-all}
\end{figure*}

\section{Representativeness of the Canonical Outcome}
\label{app:rq1-representativeness}

\begin{figure}[h]
    \centering
    \includegraphics[width=\linewidth]{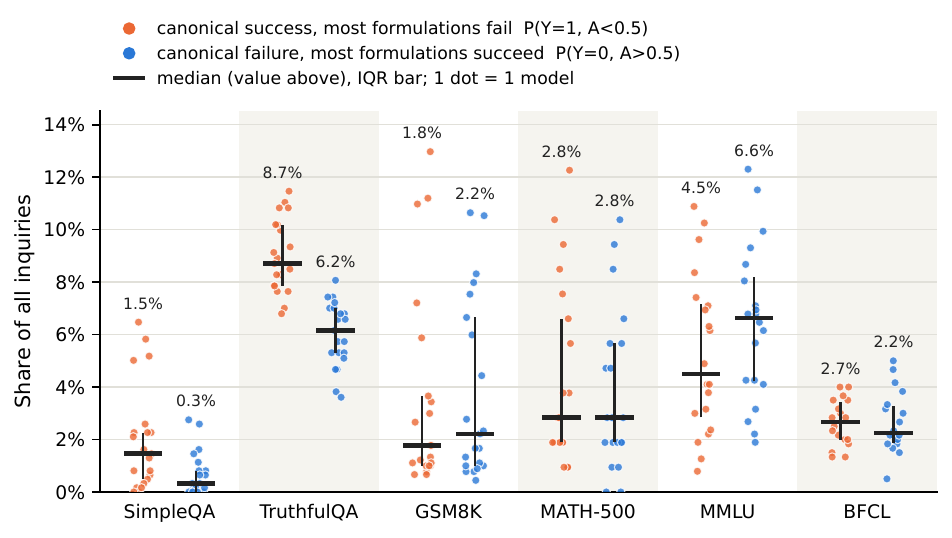}
    \caption{Majority reversals as a share of all inquiries; one dot per model,
    with the median (value above) and interquartile range.}
    \label{fig:app-reversals}
\end{figure}

Figure~\ref{fig:app-reversals} shows, for every model and benchmark, the share of
all inquiries on which the canonical outcome disagrees with the majority of
formulations, split by direction: a canonical success although most formulations
fail, $P(Y_{\mathrm{orig}}{=}1,\hat A{<}0.5)$, and a canonical failure although most
formulations succeed, $P(Y_{\mathrm{orig}}{=}0,\hat A{>}0.5)$. Both directions occur
on every benchmark, so the canonical wording neither systematically favors nor
systematically disadvantages models.

\begin{figure}[h]
    \centering
    \includegraphics[width=\linewidth]{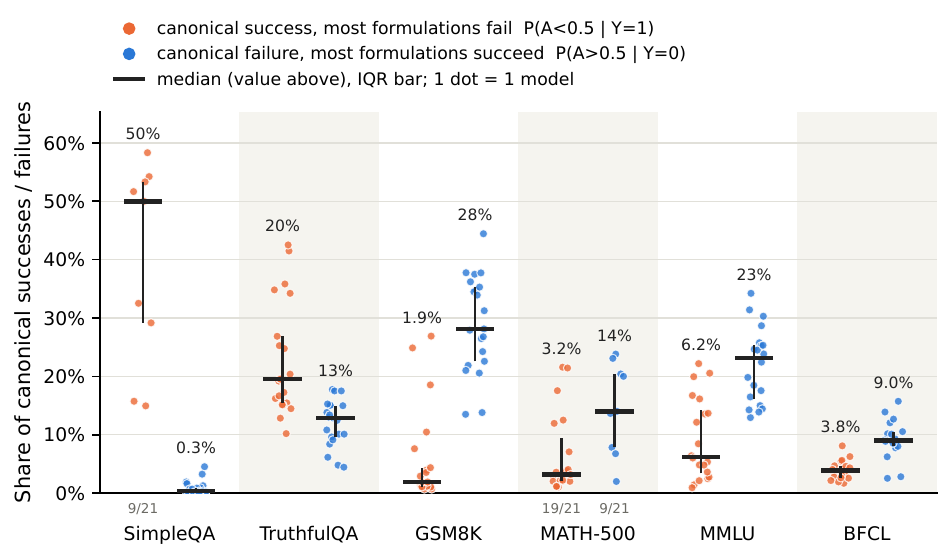}
    \caption{Majority reversals conditional on the canonical outcome,
    $P(\hat A{<}0.5\mid Y_{\mathrm{orig}}{=}1)$ and
    $P(\hat A{>}0.5\mid Y_{\mathrm{orig}}{=}0)$. A model is omitted from a group
    with fewer than 20 canonical successes (failures); counts below the axis.}
    \label{fig:app-reversals-cond}
\end{figure}

Conditioning on the canonical outcome gives the same picture from the other side
(Figure~\ref{fig:app-reversals-cond}). On GSM8K, for the median model, 64\% of
canonical failures are answered correctly under at least one alternative
formulation (41--81\% across models) and 28\% under a majority of formulations
(14--44\%), while 18\% of canonical successes fail under at least one alternative
formulation (8--82\%). The corresponding medians are 56\%, 13\%, and 65\% on
TruthfulQA and 35\%, 9\%, and 23\% on BFCL.

\section{Near-Tied Model Pairs}
\label{app:rq1-near-tied}

Across TruthfulQA, GSM8K, MATH-500, MMLU, and BFCL, 164 model pairs lie within 3
points of accuracy (median gap 1.4). For 143 (87\%) the reliable shares differ by
more than the accuracies; the median reliable-share gap is 3.8 points (95\% CI
$[3.4,4.5]$), 37\% of pairs ($[34,45]$) differ by at least 5 points, and for 89
pairs (54\%) a paired bootstrap interval excludes zero. The per-benchmark median
ranges from 3.1 (TruthfulQA, $[2.3,4.9]$) to 4.4 points (MMLU, $[2.7,6.0]$). In 40
pairs (24\%) the more accurate model has the smaller reliable share (16
significantly). In 48\% of pairs ($[40,55]$) the model with the larger reliable
share has the smaller latent share, i.e., the reliability curves cross (56\% on
BFCL, 82\% on MMLU). Among the 118 pairs whose mean accessibility differs by at
most 2 points, $\mathrm{Access}@5$ and $\mathrm{Reliable}@5$ rank the two models
in opposite orders for 62 (53\%); the disagreement is at least 2 points in both
measures for 11 pairs, 8 of them significantly in both directions, and 10 of these
11 have reliability curves that cross exactly once. The largest significant case
is LLaMA-3 8B versus LLaMA-3.1 8B on MMLU (mean accessibility within 0.1 points):
the former is 10.0 points higher in $\mathrm{Reliable}@5$ ($[6.3,13.8]$) and 9.3
points lower in $\mathrm{Access}@5$ ($[-12.8,-5.8]$). SimpleQA is excluded: most
inquiries are absent for every model, and near-tied models there share almost no
reliable mass (median reliable-share gap 0.3 points).

\begin{figure}[h]
    \centering
    \includegraphics[width=0.8\linewidth]{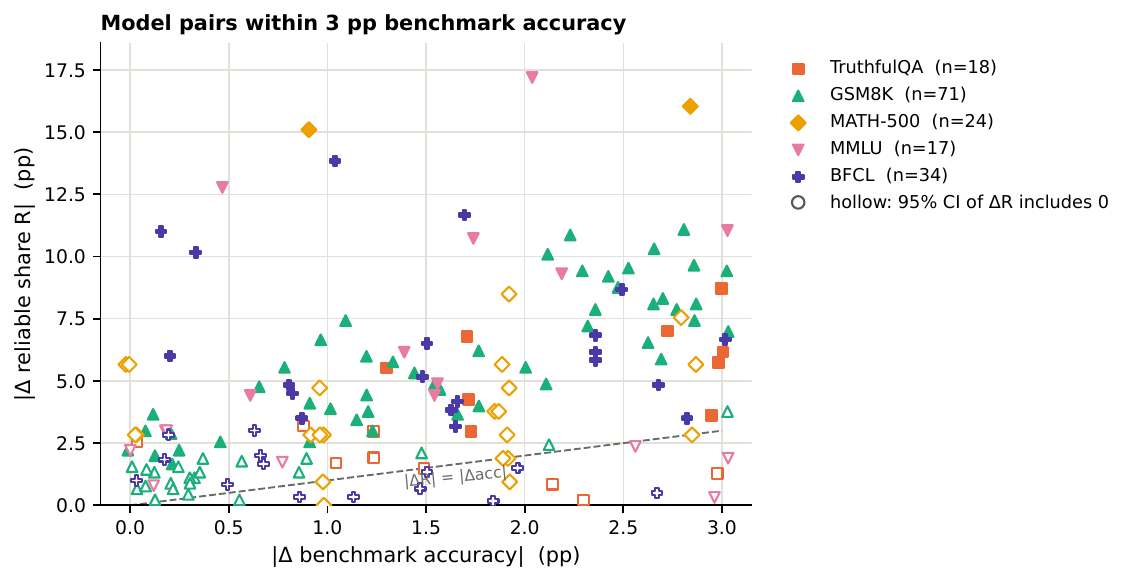}
    \caption{All model pairs within 3 points of accuracy: accuracy gap (x) versus
    reliable-share gap (y). Filled markers: paired bootstrap 95\% CI of the
    reliable-share difference excludes zero. The dashed line marks equal gaps.}
    \label{fig:app-near-tied}
\end{figure}

\section{Additional RQ2 Results}
\label{app:rq2-full}

\begin{figure*}[h]
    \centering
    \includegraphics[width=\textwidth]{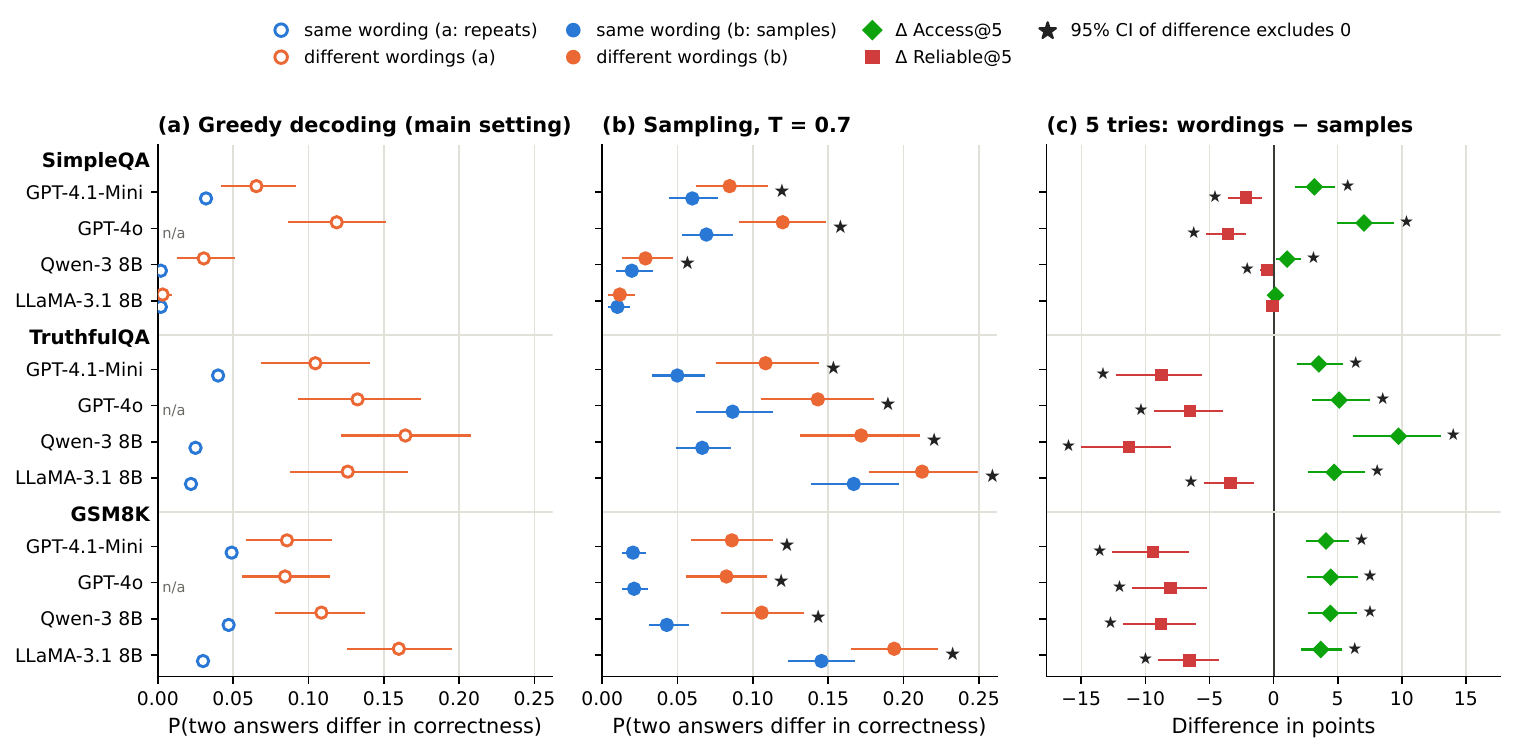}
    \caption{
    RQ2 including greedy decoding. (a)~Greedy: repeated runs of the same wording
    (blue; collected for three models on the canonical wording of a separate
    sample of 200 inquiries per benchmark; for GSM8K with a 256-token limit, which
    may inflate repeat disagreement) versus different wordings (orange).
    (b)~$T=0.7$ samples. (c)~Budget of five attempts. $\star$: the 95\% CI of the
    difference excludes zero ((b) and (c); the repeats in (a) come from a separate
    sample, so (a) has no paired CI).
    }
    \label{fig:app-rq2-greedy}
\end{figure*}

\paragraph{Greedy decoding and run-to-run variation.}
Figure~\ref{fig:app-rq2-greedy} adds the greedy comparison to the main RQ2
figure. To measure run-to-run variation under greedy decoding on the inquiries
used in the paper, we reran four models on the full GSM8K inquiry set (902
inquiries, 6{,}596 formulations) with identical settings
(Table~\ref{tab:app-rerun}). Between 1.2\% and 2.3\% of formulation outcomes
change between the two runs, about evenly in both directions and with no
systematic shift in accuracy; the generated text itself differs, reflecting
nondeterminism in the inference backends. LLaMA-3.1 8B is not part of
this baseline: its original answers were produced with a different version of the
inference software, so a rerun would not isolate run-to-run variation.

\begin{table}[h]
    \centering
    \small
    \caption{Greedy rerun of the same model on the full GSM8K inquiry set.
    ``Formulations flipped'': paraphrases whose outcome differs between the two
    runs. ``Inquiries moved'': share whose estimated accessibility differs
    (up / down).}
    \label{tab:app-rerun}
    \begin{tabular}{lccc}
    \toprule
    Model & Formulations flipped & Inquiries moved (up / down) & Canonical flips \\
    \midrule
    Qwen-3 8B & 2.2\% & 11.6\% (6.3 / 5.3) & 2.0\% \\
    Qwen-3 8B (thinking) & 1.7\% & 8.5\% (3.3 / 5.2) & 0.9\% \\
    Gemma-3 4B & 2.3\% & 11.1\% (5.3 / 5.8) & 2.4\% \\
    GPT-4.1-Mini & 1.2\% & 5.5\% (3.0 / 2.6) & 1.0\% \\
    \bottomrule
    \end{tabular}
\end{table}

\begin{figure*}[h]
    \centering
    \includegraphics[width=\textwidth]{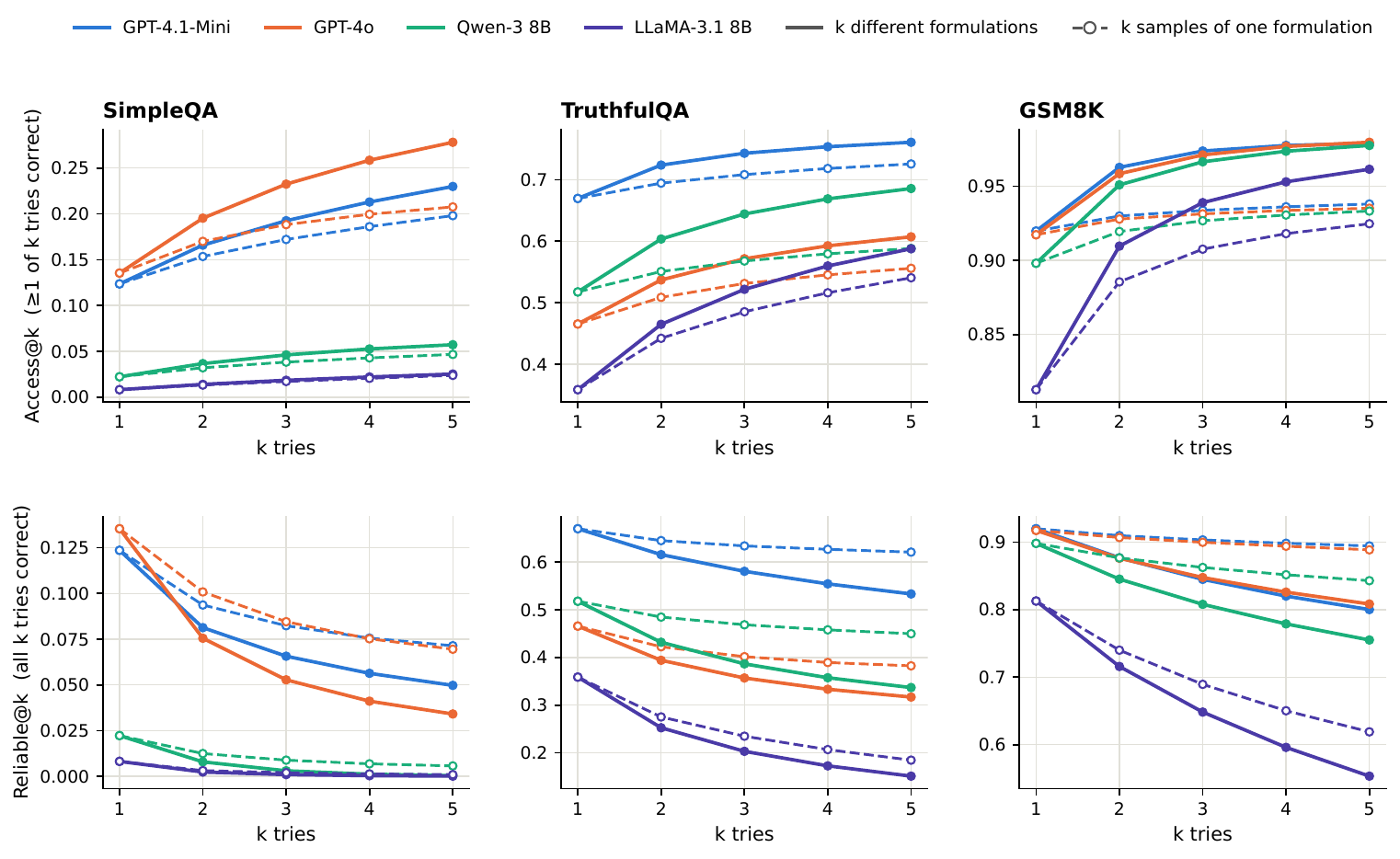}
    \caption{$\mathrm{Access}@k$ (top) and $\mathrm{Reliable}@k$ (bottom) for
    $k\le5$: solid lines distribute the $k$ attempts over $k$ different wordings,
    dashed lines draw $k$ samples of one wording ($T=0.7$).}
    \label{fig:app-rq2-k}
\end{figure*}

\section{Additional RQ3 Results}
\label{app:rq3-full}

\begin{figure*}[h]
    \centering
    \includegraphics[width=\textwidth]{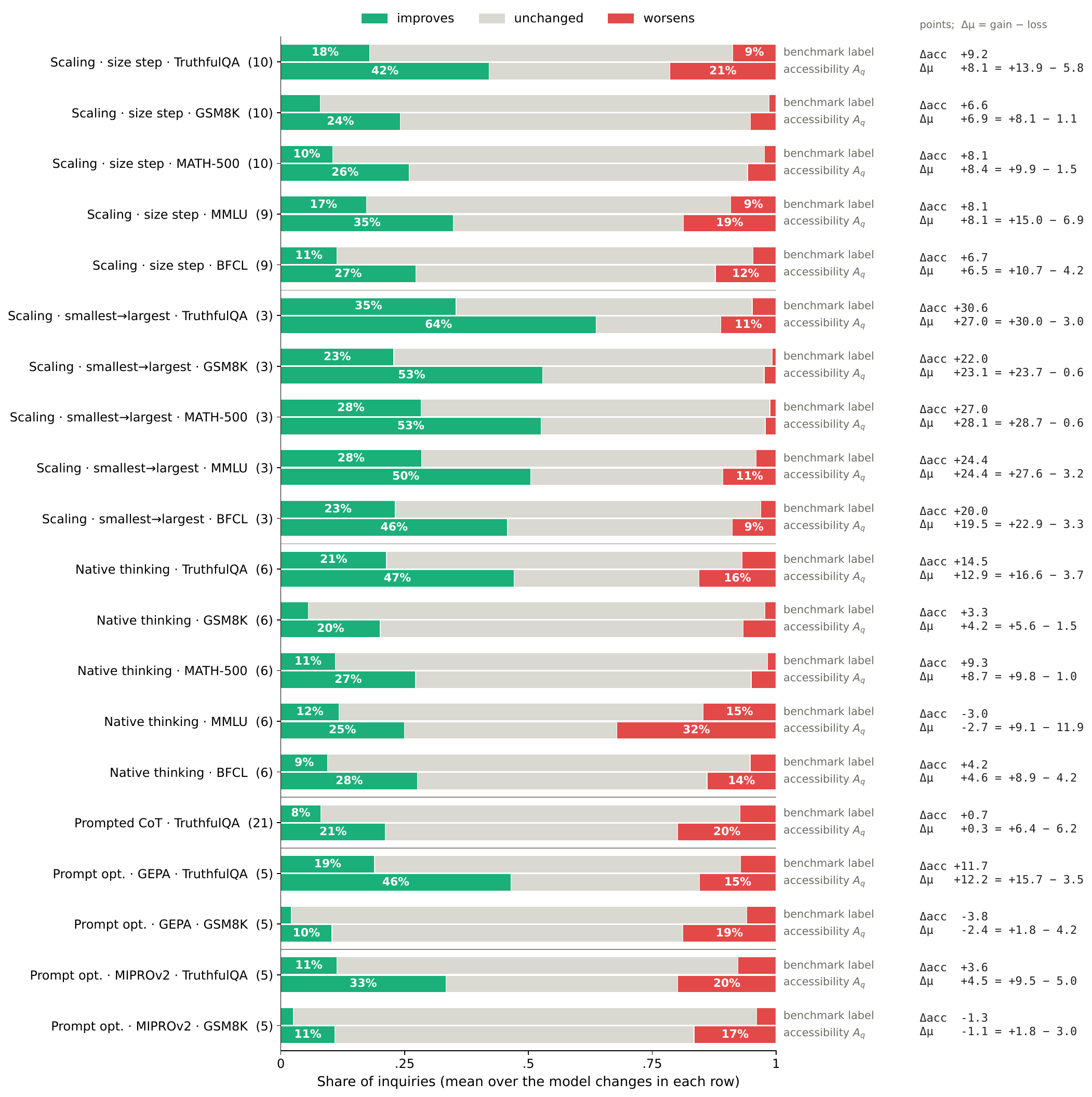}
    \caption{All intervention--benchmark rows (as in Figure~\ref{fig:rq3-two-views}),
    including smallest-to-largest scaling and GEPA and MIPROv2 separately.}
    \label{fig:app-rq3-two-views}
\end{figure*}

Most changes lie above the diagonal on GSM8K, MATH-500, and BFCL (20 of 26, 11 of
16, and 10 of 15), whereas 10 of 15 lie below it on MMLU. Among the 83 changes
that raise accuracy, the loss component of $\Delta\mu$ amounts to a median of 32\%
of the gain; among the 19 changes whose accuracy moves by at most one point, a
median of 18\% of inquiries still change accessibility (up to 74\%). Prompt
optimization on GSM8K lowers accuracy by 2.6 points on average (GEPA: 3.8) while
making 18\% of inquiries less accessible (35\% for Gemma-3 4B with GEPA).

\section{Paired Profiles for All Model Changes}
\label{app:rq3-paired}

\begin{figure*}[h]
    \centering
    \includegraphics[width=\textwidth]{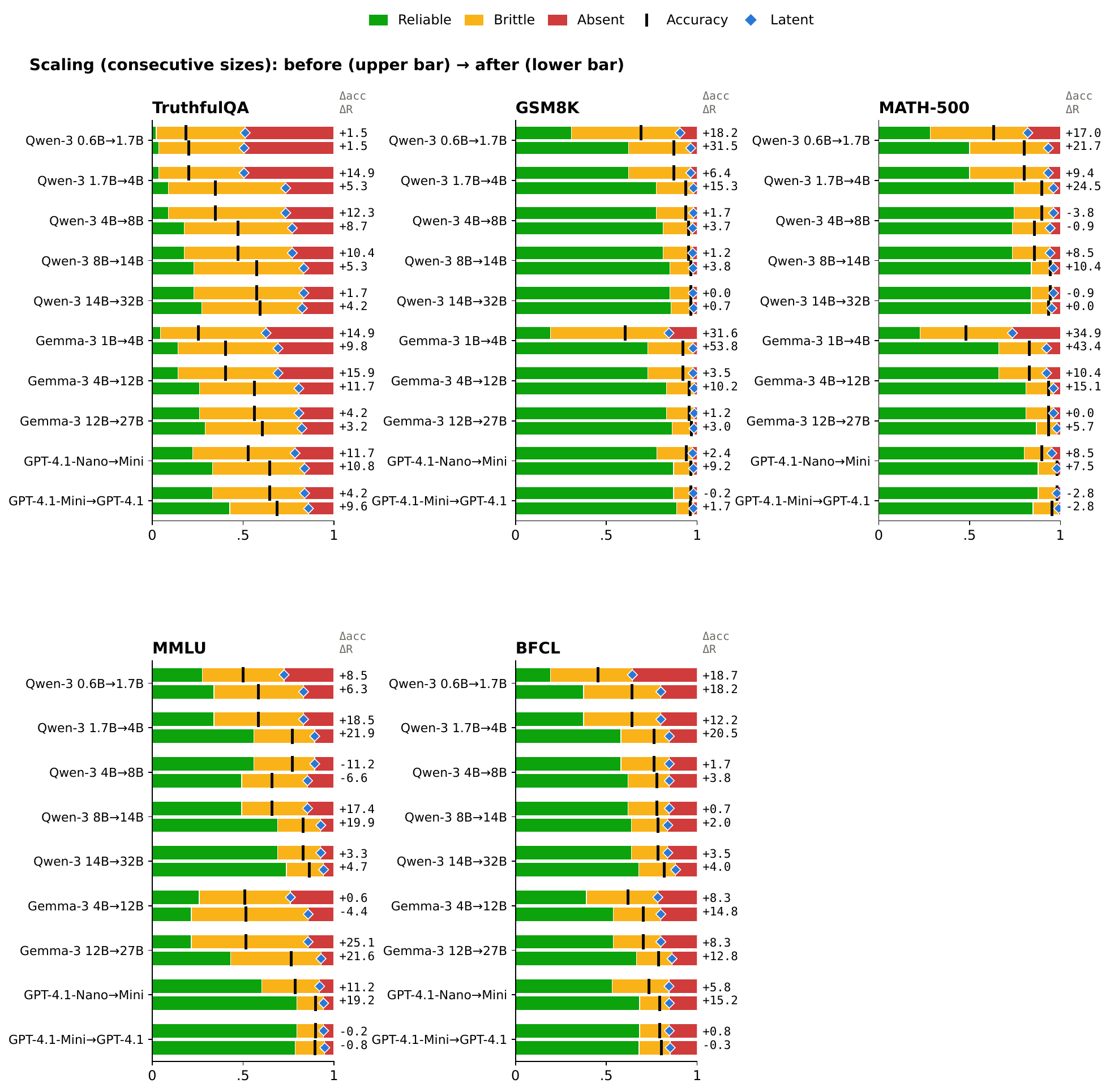}
    \caption{Paired profiles for consecutive scaling steps.}
    \label{fig:app-paired-scaling}
\end{figure*}

\begin{figure*}[h]
    \centering
    \includegraphics[width=\textwidth]{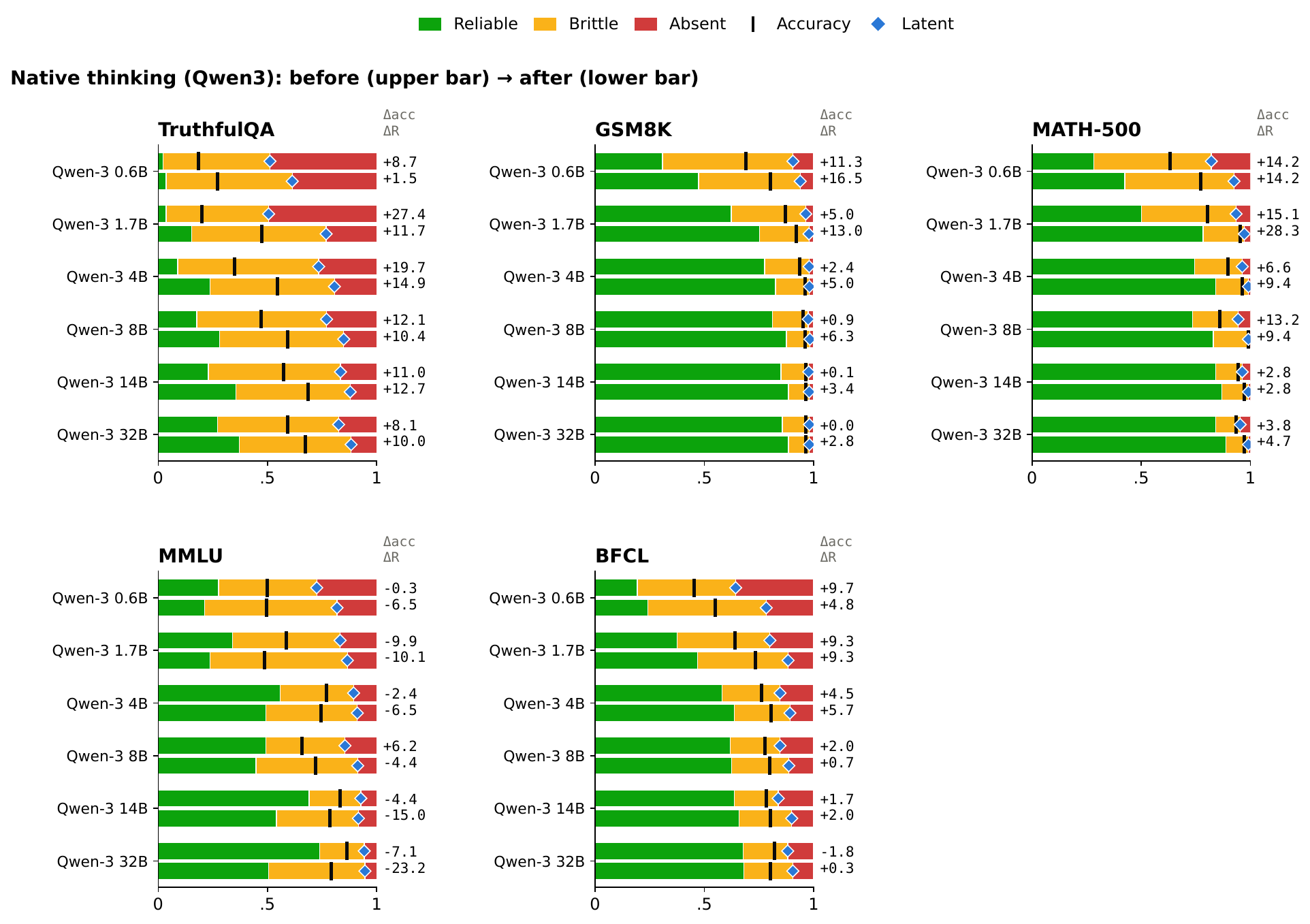}
    \caption{Paired profiles for enabling native thinking in Qwen3.}
    \label{fig:app-paired-thinking}
\end{figure*}

\begin{figure}[h]
    \centering
    \includegraphics[width=0.6\linewidth]{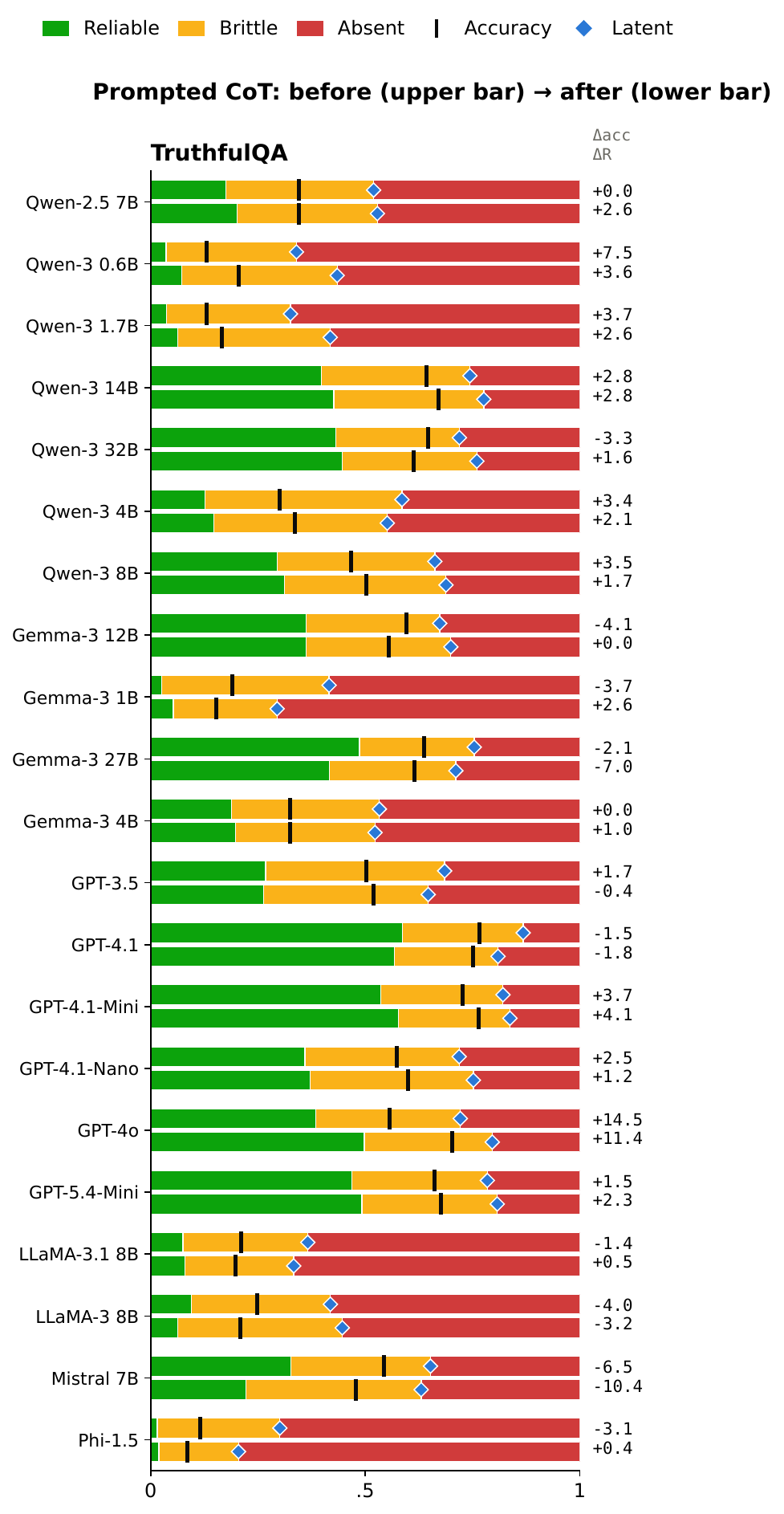}
    \caption{Paired profiles for prompted chain-of-thought on TruthfulQA.}
    \label{fig:app-paired-cot}
\end{figure}

\begin{figure*}[h]
    \centering
    \includegraphics[width=0.8\textwidth]{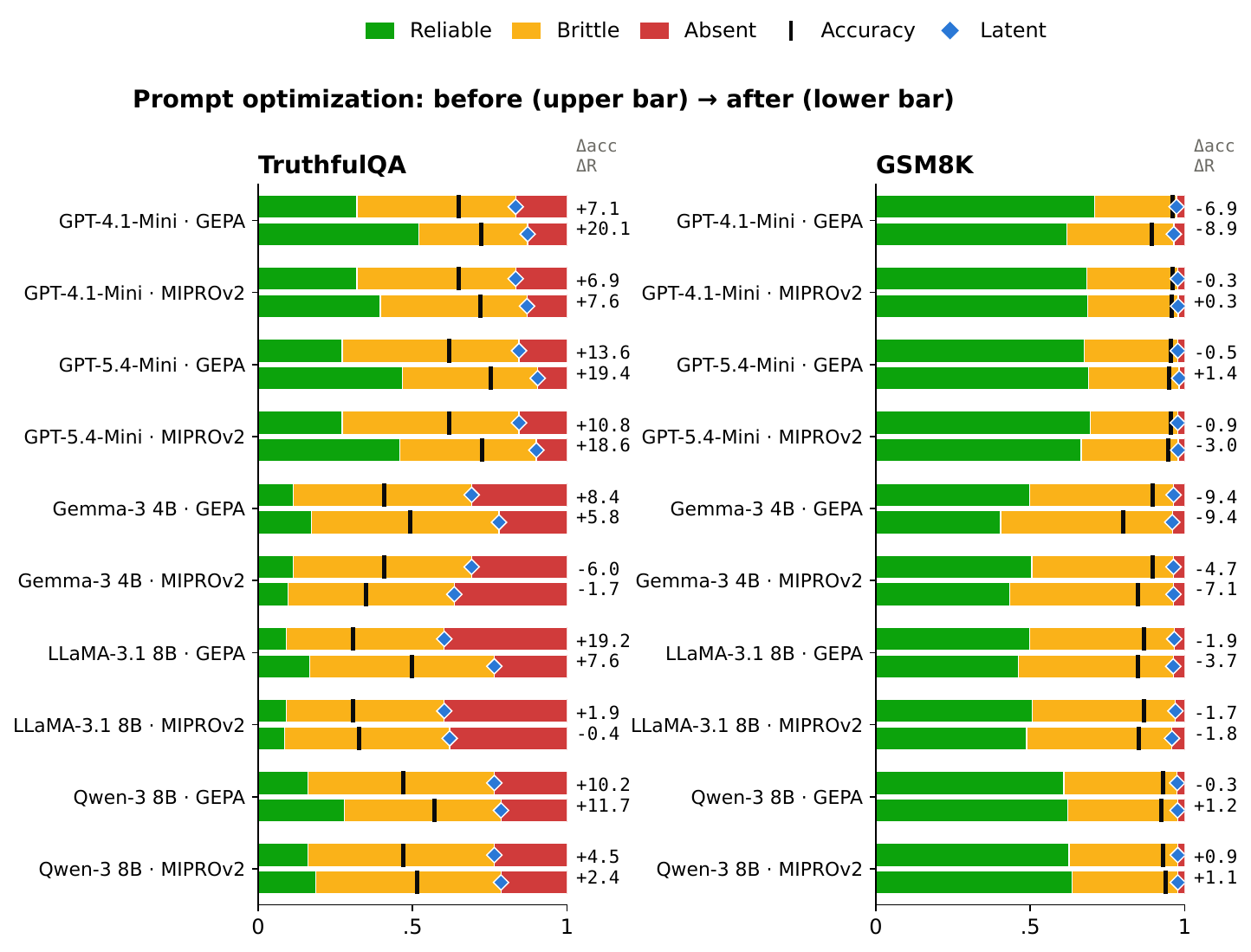}
    \caption{Paired profiles for prompt optimization with GEPA and MIPROv2.}
    \label{fig:app-paired-popt}
\end{figure*}

Table~\ref{tab:app-paired} summarizes every model change by intervention family
and benchmark, and Figures~\ref{fig:app-paired-scaling}--\ref{fig:app-paired-popt}
show the paired profiles of every individual change (before: upper bar; after:
lower bar; same encoding as Figure~\ref{fig:capability-envelope}).

\begin{table}[h]
    \centering
    \small
    \caption{Model changes by family and benchmark (means over the $n$ changes, in
    \%; before$\to$after). $A_q\!\uparrow$ / $A_q\!\downarrow$: share of inquiries whose
    accessibility increases / decreases.}
    \label{tab:app-paired}
    \resizebox{\linewidth}{!}{%
    \begin{tabular}{llr|cccccc|cc}
\toprule
Change & Benchmark & $n$ & Acc & $\mu$ & $R$ & $L$ & Access@5 & Reliable@5 & $A_q\uparrow$ & $A_q\downarrow$ \\
\midrule
Scaling & TruthfulQA & 10 & 41.7$\to$50.9 & 40.8$\to$48.9 & 15.5$\to$22.5 & 71.1$\to$76.9 & 65.2$\to$72.0 & 19.3$\to$26.8 & 42\% & 21\% \\
Scaling & GSM8K & 10 & 88.0$\to$94.6 & 85.7$\to$92.6 & 67.7$\to$81.0 & 95.7$\to$97.8 & 95.1$\to$97.6 & 71.3$\to$83.5 & 24\% & 5\% \\
Scaling & MATH-500 & 10 & 82.5$\to$90.7 & 80.7$\to$89.1 & 64.8$\to$77.3 & 91.8$\to$95.9 & 90.6$\to$95.4 & 67.8$\to$79.8 & 26\% & 6\% \\
Scaling & MMLU & 9 & 67.2$\to$75.3 & 67.7$\to$75.9 & 47.0$\to$56.1 & 85.8$\to$90.4 & 83.5$\to$88.7 & 50.1$\to$59.5 & 35\% & 19\% \\
Scaling & BFCL & 9 & 69.7$\to$76.3 & 68.3$\to$74.8 & 50.5$\to$60.6 & 80.6$\to$84.1 & 78.2$\to$82.3 & 55.9$\to$65.1 & 27\% & 12\% \\
Thinking & TruthfulQA & 6 & 39.5$\to$54.0 & 39.3$\to$52.2 & 13.7$\to$23.9 & 69.7$\to$80.0 & 63.8$\to$75.4 & 17.7$\to$28.8 & 47\% & 16\% \\
Thinking & GSM8K & 6 & 89.6$\to$92.9 & 87.5$\to$91.6 & 70.5$\to$78.3 & 96.4$\to$97.4 & 95.9$\to$97.1 & 73.9$\to$81.4 & 20\% & 7\% \\
Thinking & MATH-500 & 6 & 84.4$\to$93.7 & 82.7$\to$91.4 & 65.7$\to$77.2 & 92.9$\to$97.6 & 92.2$\to$97.3 & 69.1$\to$80.7 & 27\% & 5\% \\
Thinking & MMLU & 6 & 70.1$\to$67.1 & 70.7$\to$67.9 & 51.6$\to$40.6 & 86.3$\to$89.5 & 84.4$\to$87.2 & 54.7$\to$44.9 & 25\% & 32\% \\
Thinking & BFCL & 6 & 70.6$\to$74.9 & 68.9$\to$73.6 & 51.4$\to$55.2 & 80.9$\to$87.5 & 78.5$\to$84.6 & 56.7$\to$60.6 & 28\% & 14\% \\
CoT & TruthfulQA & 21 & 44.4$\to$45.1 & 42.8$\to$43.1 & 27.0$\to$27.9 & 60.1$\to$60.2 & 57.2$\to$57.4 & 29.4$\to$30.2 & 21\% & 20\% \\
Prompt opt. & TruthfulQA & 10 & 49.0$\to$56.7 & 46.2$\to$54.6 & 19.2$\to$28.3 & 74.7$\to$79.2 & 68.8$\to$74.5 & 24.4$\to$33.8 & 40\% & 18\% \\
Prompt opt. & GSM8K & 10 & 92.2$\to$89.6 & 86.0$\to$84.2 & 60.0$\to$57.0 & 97.2$\to$97.0 & 96.4$\to$95.9 & 67.6$\to$64.6 & 11\% & 18\% \\
\bottomrule
\end{tabular}
    }
\end{table}

\section{Estimating Profiles Under a Fixed Budget}
\label{app:budget}

Evaluating $m$ formulations for every inquiry costs $m$ times a conventional
run. We ask whether the profile can instead be estimated at the cost of a single
run of $N$ calls, by evaluating $K$ random formulations on $n=N/K$ random
inquiries. For each benchmark and model we draw 2{,}000 such subsamples from the
paper's data (for each inquiry, the number of successes among $K$ of its $m$
formulations is hypergeometric) and record how often each estimate falls within
5 points of its full-data value: mean accessibility, and the latent and reliable
shares over $K$ formulations ($\mathrm{Access}@K$ and $\mathrm{Reliable}@K$).
For comparison, we apply the same criterion to accuracy on the same $n$
inquiries. Table~\ref{tab:app-budget} reports the share of models for which an
estimate meets this tolerance in at least 95\% of subsamples.

With two formulations on half of the inquiries, every quantity meets the
tolerance for every model on SimpleQA, GSM8K, MMLU, and BFCL, exactly as accuracy
does, and mean accessibility is as recoverable as accuracy on every benchmark
except the small MATH-500. When fewer inquiries remain, because $K$ is larger or
the benchmark is small (TruthfulQA, MATH-500), accuracy is not recovered this
precisely either, and the reliable share is the hardest quantity to estimate
(e.g., MMLU and BFCL at $K=3$). In practice, a benchmark with a few hundred
inquiries supports estimating the profile with two formulations per inquiry at
the cost of a single conventional run.

\begin{table}[h]
    \centering
    \small
    \caption{Share of models whose estimate is within 5 points of the full-data
    value in at least 95\% of equal-cost subsamples ($n=N/K$ inquiries, $K$
    formulations each). Acc.: accuracy on the same $n$ inquiries.}
    \label{tab:app-budget}
    \begin{tabular}{lrr|cccc}
    \toprule
    Benchmark & $K$ & $n$ & Acc. & $\mu$ & $\mathrm{Access}@K$ & $\mathrm{Reliable}@K$ \\
    \midrule
    SimpleQA   & 2 & 309 & 1.00 & 1.00 & 1.00 & 1.00 \\
               & 3 & 206 & 0.95 & 0.95 & 0.95 & 1.00 \\
    TruthfulQA & 2 & 235 & 1.00 & 1.00 & 0.48 & 0.52 \\
               & 3 & 157 & 0.00 & 0.29 & 0.00 & 0.29 \\
    GSM8K      & 2 & 451 & 1.00 & 1.00 & 1.00 & 1.00 \\
               & 3 & 300 & 1.00 & 1.00 & 1.00 & 0.90 \\
    MATH-500   & 2 &  53 & 0.43 & 0.29 & 0.19 & 0.05 \\
    MMLU       & 2 & 317 & 1.00 & 1.00 & 1.00 & 1.00 \\
               & 3 & 211 & 0.55 & 1.00 & 0.60 & 0.30 \\
    BFCL       & 2 & 300 & 1.00 & 1.00 & 1.00 & 1.00 \\
               & 3 & 200 & 0.50 & 0.67 & 0.61 & 0.06 \\
    \bottomrule
    \end{tabular}
\end{table}


\section{Breadth--Repeatability Decomposition of Mean Accessibility}
\label{app:breadth-repeatability}

Mean accessibility $\mu=\mathbb{E}_q[A_q]$ does not distinguish whether success
is spread moderately across many inquiries or concentrated on a smaller subset
of inquiries on which it is highly accessible. This appendix gives a continuous
decomposition of $\mu$ that separates these two cases and applies it to the
model changes studied in Section~\ref{sec:rq3}.

\paragraph{Definition.}
For a profile with $\mu>0$, define
\begin{equation}
D=\frac{\mathbb{E}_q[A_q^2]}{\mathbb{E}_q[A_q]},
\qquad
B=\frac{\mathbb{E}_q[A_q]^2}{\mathbb{E}_q[A_q^2]},
\qquad\text{so that}\qquad
\mu=B\,D .
\label{eq:breadth-repeatability}
\end{equation}

\paragraph{Repeatability $D$.}
Sample an inquiry $q$ uniformly and two formulations $f_1,f_2\sim D_q$
independently, and let $S_i$ denote success on $f_i$. Given $q$, $S_1$ and $S_2$
are independent with probability $A_q$ each, so
\[
\Pr(S_2\mid S_1)=\frac{\Pr(S_1,S_2)}{\Pr(S_1)}
=\frac{\mathbb{E}_q[A_q^2]}{\mathbb{E}_q[A_q]}=D .
\]
$D$ is therefore the probability that a success is repeated under a fresh
formulation of the same inquiry. Equivalently, $D$ is the mean of $A$ over
inquiries weighted by $A$ (the accessibility of the inquiry behind a randomly
chosen success), and $D=\mu+\operatorname{Var}_q(A)/\mu$.

\paragraph{Effective breadth $B$.}
Let $p_q=A_q/\sum_{q'}A_{q'}$ be the distribution of accessibility mass over the
$n$ inquiries. Then
\[
nB=\frac{\bigl(\sum_q A_q\bigr)^2}{\sum_q A_q^2}=\frac{1}{\sum_q p_q^2},
\]
the inverse Simpson index (effective number) of this distribution; $B$ is the
corresponding evenness. If $A_q=c$ on $k$ inquiries and $0$ elsewhere, then
$nB=k$ and $D=c$ exactly.

\paragraph{Properties.}
\begin{itemize}
\item $\mu\le D\le 1$: $A^2\le A$ on $[0,1]$ gives $D\le1$, and
  $\mathbb{E}[A^2]\ge\mathbb{E}[A]^2$ gives $D\ge\mu$.
\item $\mu\le B\le P(A>0)$: $B=\mu/D\ge\mu$, and by Cauchy--Schwarz
  $\mathbb{E}[A]^2=\mathbb{E}[A\,\mathbf{1}\{A>0\}]^2\le\mathbb{E}[A^2]\,P(A>0)$,
  with equality iff $A$ is constant on its support. Effective breadth never
  exceeds the latent share.
\item With $\mathrm{CV}$ the coefficient of variation of $A$ across inquiries,
  $B=1/(1+\mathrm{CV}^2)$ and $D=\mu\,(1+\mathrm{CV}^2)$. At fixed $\mu$, the
  split between $B$ and $D$ is thus governed by a single quantity: how unevenly
  accessibility is distributed across inquiries. $B$ is invariant to rescaling
  $A$, whereas $D$ scales with it, so $D$ partly tracks $\mu$.
\item Changes decompose additively in logs:
  $\Delta\log\mu=\Delta\log B+\Delta\log D$.
\end{itemize}

\begin{figure*}[t]
    \centering
    \includegraphics[width=\textwidth]{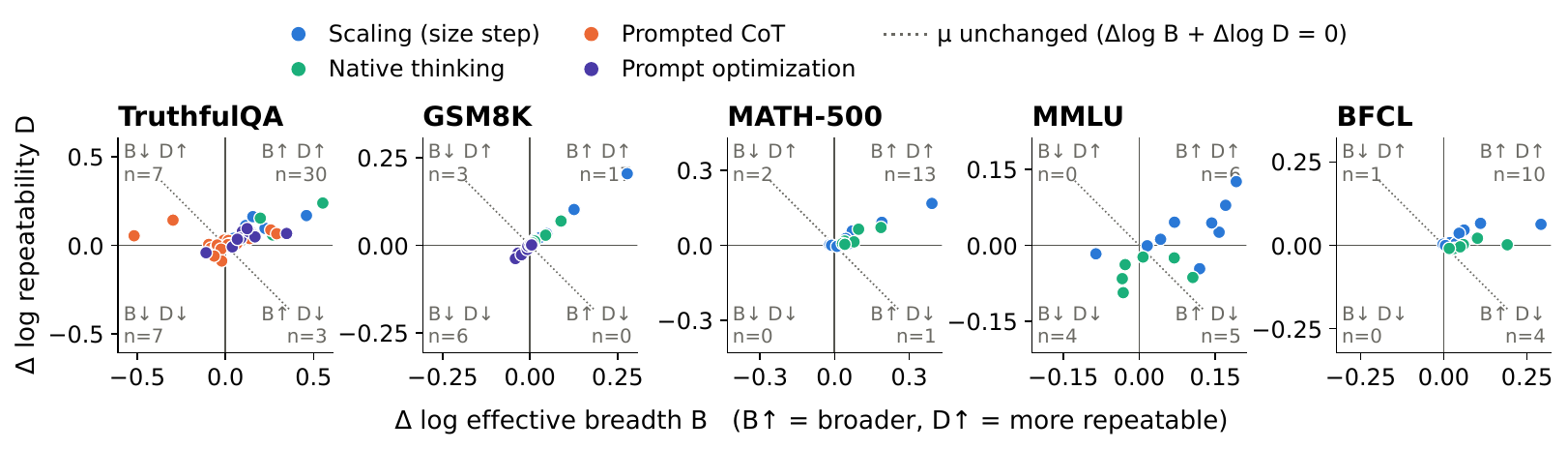}
    \caption{
    \textbf{Breadth--repeatability decomposition of model changes.}
    Each point is one model change; $\Delta\log B$ (x) and $\Delta\log D$ (y)
    sum to $\Delta\log\mu$, so the dotted line marks unchanged mean
    accessibility. Corner labels give the number of changes in each quadrant.
    }
    \label{fig:app-breadth-repeatability}
\end{figure*}

\paragraph{Estimation.}
With $k$ successes among $m$ formulations, we estimate $\mathbb{E}_q[A^2]$ by
the plug-in mean of $\hat A_q^2=(k/m)^2$. Because
$\mathbb{E}[\hat A^2\mid q]=A^2+A(1-A)/m$, the plug-in estimate slightly
inflates $D$ and deflates $B$; it corresponds to drawing the second formulation
from the same $m$ formulations with replacement. The unbiased alternative
$k(k-1)/(m(m-1))$ corresponds to requiring a \emph{different} second
formulation. On our data the two estimators place 118 of the 119 model changes
below in the same quadrant of Figure~\ref{fig:app-breadth-repeatability}.

\paragraph{Results.}
Figure~\ref{fig:app-breadth-repeatability} places each model change from
Section~\ref{sec:rq3} by $\Delta\log B$ and $\Delta\log D$. For most beneficial
scaling and reasoning changes, breadth and repeatability increase together: 36
of 46 consecutive scaling steps, and all 18 native-thinking comparisons on
TruthfulQA, GSM8K, and MATH-500. The exceptions are informative. On MMLU, native
thinking lowers repeatability for all six Qwen3 sizes; for three of them
effective breadth increases at the same time, so successful behavior is spread
across more inquiries but is less reproducible across formulations. The same
broader-but-less-repeatable pattern appears for three of six sizes on BFCL.
Prompt optimization on GSM8K tends instead to reduce both quantities (6 of 10
changes across GEPA and MIPROv2), whereas on TruthfulQA it mostly increases both
(8 of 10). Prompted CoT on TruthfulQA spans all four regimes (8, 2, 5, and 6 of
21 changes in the broader/more-repeatable, broader/less-repeatable,
narrower/more-repeatable, and narrower/less-repeatable quadrants).

\end{document}